\PassOptionsToPackage{table}{xcolor}

\documentclass[sigconf]{acmart}

\copyrightyear{2024}
\acmYear{2024}
\setcopyright{acmlicensed}
\acmConference[KDD '24]{Proceedings of the 30th ACM SIGKDD Conference on Knowledge Discovery and Data Mining}{August 25--29, 2024}{Barcelona, Spain}
\acmBooktitle{Proceedings of the 30th ACM SIGKDD Conference on Knowledge Discovery and Data Mining (KDD '24), August 25--29, 2024, Barcelona, Spain}
\acmDOI{10.1145/3637528.3671900}
\acmISBN{979-8-4007-0490-1/24/08}

\author{Pranjal Aggarwal}
\authornote{Equal Contribution}
\affiliation{%
  \institution{Indian Institute of Technology Delhi}
  \city{New Delhi}
  \country{India}
  }
 \email{pranjal2041@gmail.com}

\author{Vishvak Murahari}
\authornotemark[1]
\affiliation{%
  \institution{Princeton University}
  \city{Princeton}
  \country{USA}
  }
 \email{murahari@cs.princeton.edu}

\author{Tanmay Rajpurohit}
\affiliation{%
  \institution{Independent}
  \city{Seattle}
  \country{USA}
  }
  \email{tanmay.rajpurohit@gmail.com}

\author{Ashwin Kalyan}
\affiliation{%
  \institution{Independent}
  \city{Seattle}
  \country{USA}
  }
  \email{asaavashwin@gmail.com}
  
\author{Karthik Narasimhan}
\affiliation{%
  \institution{Princeton University}
  \city{Princeton}
  \country{USA}
  }
  \email{karthikn@princeton.edu}
  
\author{Ameet Deshpande}
\affiliation{%
  \institution{Princeton University}
  \city{Princeton}
  \country{USA}
  }
\email{asd@princeton.edu}

\renewcommand{\shortauthors}{Pranjal Aggarwal et al.}

\definecolor{mydarkblue}{rgb}{0,0.08,0.45}

\usepackage{hyperref}
\hypersetup{
    colorlinks=true,
    linkcolor=blue,
    filecolor=magenta,
    urlcolor=cyan,
    citecolor=mydarkblue,
}
\usepackage[normalem]{ulem}
\usepackage{xspace}
\usepackage{booktabs,arydshln}
\usepackage{multirow}
\usepackage{caption}
\usepackage{newfloat}
\usepackage{enumitem} %

\usepackage{amsmath,amsfonts,bm}

\def\eqref#1{equation~\ref{#1}}

\def\1{\bm{1}}

\DeclareMathAlphabet{\mathsfit}{\encodingdefault}{\sfdefault}{m}{sl}
\SetMathAlphabet{\mathsfit}{bold}{\encodingdefault}{\sfdefault}{bx}{n}

\usepackage{listings}

\lstdefinelanguage{markdown}{
  basicstyle=\ttfamily\footnotesize,
  columns=fullflexible,
  breaklines=true,
  frame=single,
  backgroundcolor=\color{gray!10},
  morekeywords={Write, Question, Search, Results},
  morecomment=[l]{//},
  morecomment=[s]{/*}{*/},
  morestring=[b]",
  morestring=[b]',
  moredelim=[s][\color{blue}\bfseries]{[}{]},
  moredelim=[s][\color{red}]{(U+}{)},
}

\usepackage{hyperref}
\usepackage{url}

\usepackage{graphicx}
\usepackage{epstopdf}
\usepackage{algorithm}
\usepackage{algorithmicx}
\usepackage{algpseudocode}
\usepackage{subfig}

\DeclareFloatingEnvironment[name=Listing,placement=ht,fileext=los]{codeblock}

\newcommand{\todo}[1]{\textcolor{red}{[\textcolor{blue}{TODO:} ]}}

\newcommand{\cmmnt}[1]{\ignorespaces}

\newcommand{\LE}{GE}        
\newcommand{\lee}{generative engine}
\newcommand{\Le}{Generative Engine}     
\newcommand{\llm}{LLM}

\newcommand{\leo}{\textsc{Generative Engine Optimization}}
\newcommand{\LEO}{\textsc{GEO}}

\newcommand{\bench}{\textsc{GEO-bench}}

\newcommand{\leobaseline}{No Optimization}
\newcommand{\leoseo}{Keyword Stuffing}
\newcommand{\leosimple}{Easy-to-Understand}
\newcommand{\leoauthoritative}{Authoritative}
\newcommand{\leounique}{Unique Words}
\newcommand{\leotechnical}{Technical Terms}
\newcommand{\leofluent}{Fluency Optimization}
\newcommand{\leociting}{Cite Sources}
\newcommand{\leoquotes}{Quotation Addition}
\newcommand{\leostats}{Statistics Addition}

\newcommand{\wordposmetric}{Position-Adjusted Word Count}
\newcommand{\subjectiveimpression}{Subjective Impression}

\definecolor{redback}{HTML}{ff957e}
\definecolor{greenback}{HTML}{8ee0b6}

\newcommand{\tablestd}[1]{{\tiny \textcolor{gray}{$(\pm #1)$}}}

\begin{document}

\title{GEO: Generative Engine Optimization}



\begin{abstract}

The advent of large language models (LLMs) has ushered in a new paradigm of search engines that use generative models to gather and summarize information to answer user queries.
This emerging technology, which we formalize under the unified framework of \lee{}s (GEs), can generate accurate and personalized responses, rapidly replacing traditional search engines like Google and Bing.
\Le{}s typically satisfy queries by synthesizing information from multiple sources and summarizing them using LLMs.
While this shift significantly improves \textit{user} utility and \textit{generative search engine} traffic, it poses a huge challenge for the third stakeholder -- website and content creators.
Given the black-box and fast-moving nature of \lee{}s, content creators have little to no control over \textit{when} and \textit{how} their content is displayed.
With \lee{}s here to stay, we must ensure the creator economy is not disadvantaged.
To address this, we introduce \leo{} (\LEO{}), the first novel paradigm to aid content creators in improving their content visibility in \lee{} responses through a flexible black-box optimization framework for optimizing and defining visibility metrics.
We facilitate systematic evaluation by introducing \bench{}, a large-scale benchmark of diverse user queries across multiple domains, along with relevant web sources to answer these queries.
Through rigorous evaluation, we demonstrate that \LEO{} can boost visibility by up to 40\% in \lee{} responses.
Moreover, we show the efficacy of these strategies varies across domains, underscoring the need for domain-specific optimization methods.
Our work opens a new frontier in information discovery systems, with profound implications for both developers of \lee{}s and content creators.\footnote{Code and Data available at \url{https://generative-engines.com/GEO/}}

\end{abstract}

\maketitle

\section{Introduction}
\label{sec:intro}

\begin{figure*}[t]
    \centering
    \includegraphics[width=0.99\linewidth]{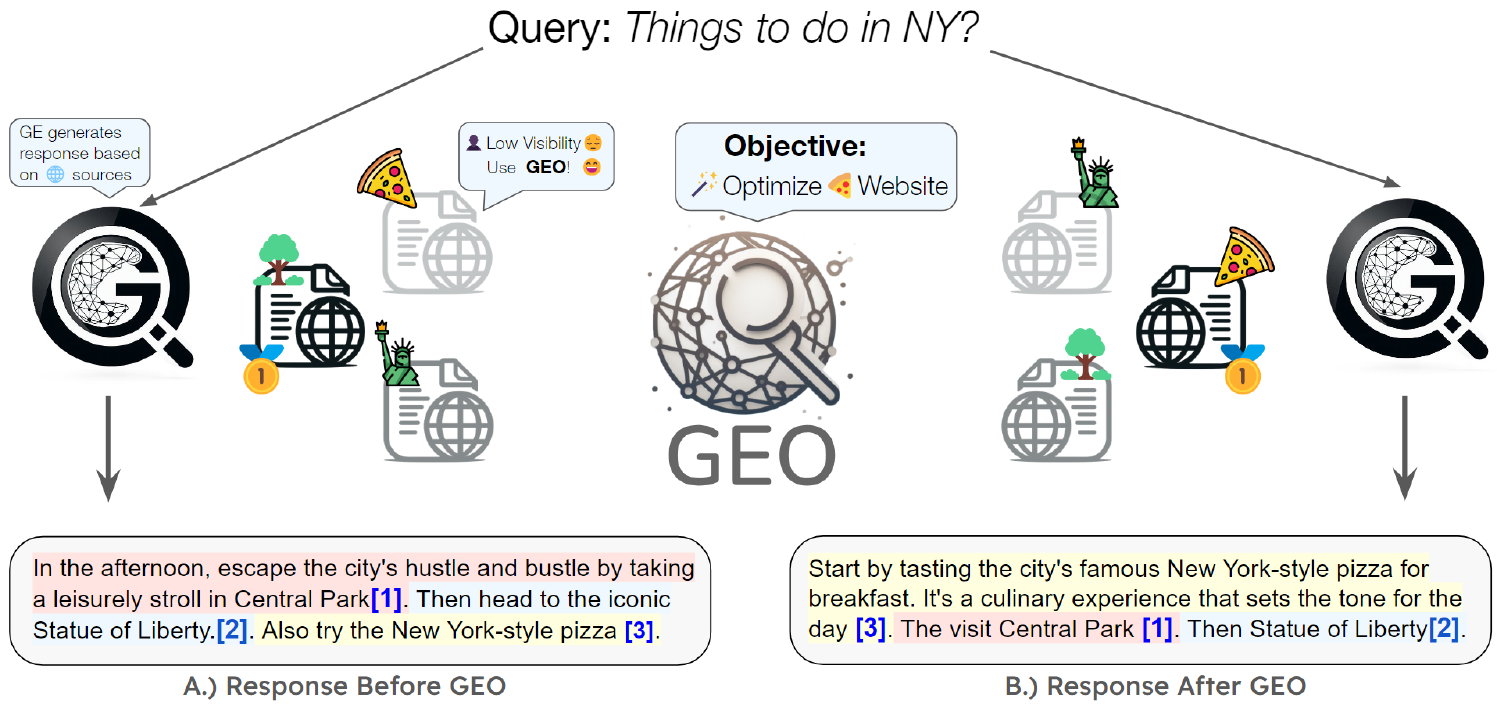}
    \caption{
    Our proposed \leo{} (\LEO{}) method optimizes websites to boost their visibility in \Le{} responses.
    \LEO{}'s black-box optimization framework then enables the website owner of the pizza website, which lacked visibility originally, to optimize their website to increase visibility under \Le{}s.
    Further, \LEO{}'s general framework allows content creators to define and optimize their custom visibility metrics, giving them greater control in this new emerging paradigm.
    }
    \label{fig:leo_teaser}
\end{figure*}

The invention of traditional search engines three decades ago revolutionized information access and dissemination globally~\cite{Brin1998TheAO}. While they were powerful and ushered in a host of applications like academic research and e-commerce, they were limited to providing a list of relevant websites for user queries. However, the recent success of large language models~\cite{NEURIPS2020_1457c0d6, openai2024gpt4} has paved the way for better systems like BingChat, Google's SGE, and perplexity.ai that combine conventional search engines with generative models. We dub these systems \lee{}s (\LE{}) because they \textit{search} for information and \textit{generate} multi-modal responses by using multiple sources. Technically, \lee{}s (Figure~\ref{fig:le_overview}) retrieve relevant documents from a database (like the internet) and use large neural models to generate a response grounded on the sources, ensuring attribution and a way for the user to verify the information.

The usefulness of \lee{}s for developers and users is evident -- users access information faster and more accurately,
while developers craft precise and personalized responses, improving user satisfaction and revenue. However, \lee{}s disadvantage the third stakeholder -- website and content creators. \Le{}s, in contrast to traditional search engines, remove the need to navigate to websites by directly providing a precise and comprehensive response, potentially reducing organic traffic to websites and impacting their visibility~\cite{Maayan2023}. With millions of small businesses and individuals relying on online traffic and visibility for their livelihood, \lee{}s will significantly disrupt the creator economy.
Further, the black-box and proprietary nature of \lee{}s makes it difficult for content creators to \textit{control} and \textit{understand} how their content is ingested and portrayed.

In this work, we propose the first general creator-centric framework to optimize content for \lee{}s, which we dub \leo{} (\LEO{}), to empower content creators to navigate this new search paradigm.
\LEO{} is a flexible black-box optimization framework for optimizing web content visibility for proprietary and closed-source generative engines (Figure~\ref{fig:leo_teaser}). \LEO{} ingests a source website and outputs an optimized version by tailoring and calibrating the presentation, text style, and content to increase visibility in \lee{}s.

Further, \LEO{} introduces a flexible framework for defining visibility metrics tailor-made for \lee{}s as the notion of visibility in \lee{}s is more nuanced and multi-faceted than traditional search engines (Figure~\ref{fig:geo_explain}).
While average ranking on the response page is a good measure of visibility in traditional search engines, which present a linear list of websites, this does not apply to \lee{}s.
\Le{}s provide rich, structured responses and embed websites as inline citations in the response, often embedding them with different lengths, at varying positions, and with diverse styles.
This necessitates the need for visibility metrics tailor-made for \lee{}s, which measure the visibility of attributed sources over multiple dimensions, such as relevance and influence of citation to query, measured through both an objective and a subjective lens.

To facilitate faithful and extensive evaluation of \LEO{} methods, we propose \bench{}, a benchmark consisting of 10000 queries from diverse domains and sources, adapted for \lee{}s. 
Through systematic evaluation, we demonstrate that our proposed \leo{} methods can boost visibility by up to 40\% on diverse queries, providing beneficial strategies for content creators. Among other things, we find that including citations, quotations from relevant sources, and statistics can significantly boost source visibility, with an increase of over 40\% across various queries. We also demonstrate the efficacy of \leo{} on Perplexity.ai, a real-world \lee{} and demonstrate visibility improvements up to 37\%.

In summary, our contributions are three-fold:\\
(1) We propose \leo{}, the first general optimization framework for website owners to optimize their websites for \lee{}s. \leo{} can improve the visibility of websites by up to 40\% on a wide range of queries, domains, and real-world black-box generative engines. \\
(2) Our framework proposes a comprehensive set of visibility metrics specifically designed for \lee{}s and enables content creators to flexibly optimize their content through customized visibility metrics. \\
(3) To foster faithful evaluation of \LEO{} methods in \lee{}s, we propose the first large-scale benchmark consisting of diverse search queries from wide-ranging domains and datasets specially tailored for \Le{}s.

\section{Formulation \& Methodology}

\subsection{Formulation of \Le{}s}
\label{sec:formulation_le}

\begin{figure}[t]
    \centering
    \includegraphics[width=\linewidth]{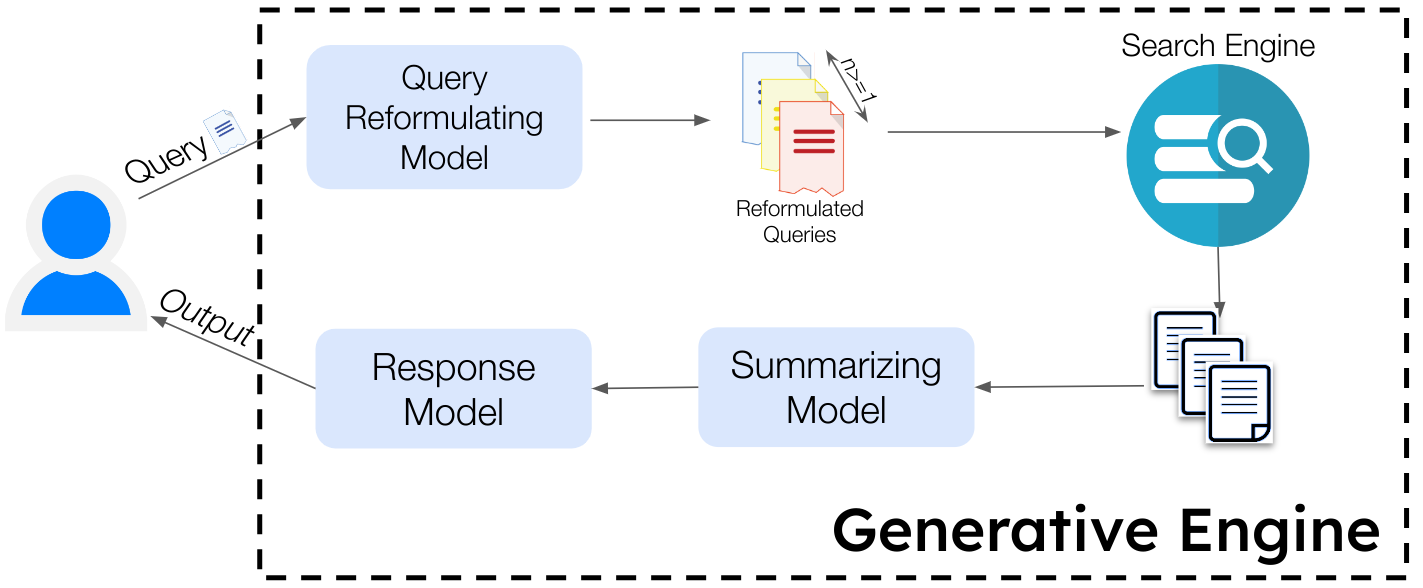}
    \caption{Overview of \Le{}s. \Le{}s primrarily consists of a set of generative models and a search engine to retrieve relevant documents. \Le{}s take user query as input and through a series of steps generate a final response that is grounded in the retrieved sources with inline attributions.}
    \label{fig:le_overview}
\end{figure}
\begin{figure*}[t]
    \centering
    \includegraphics[width=\linewidth]{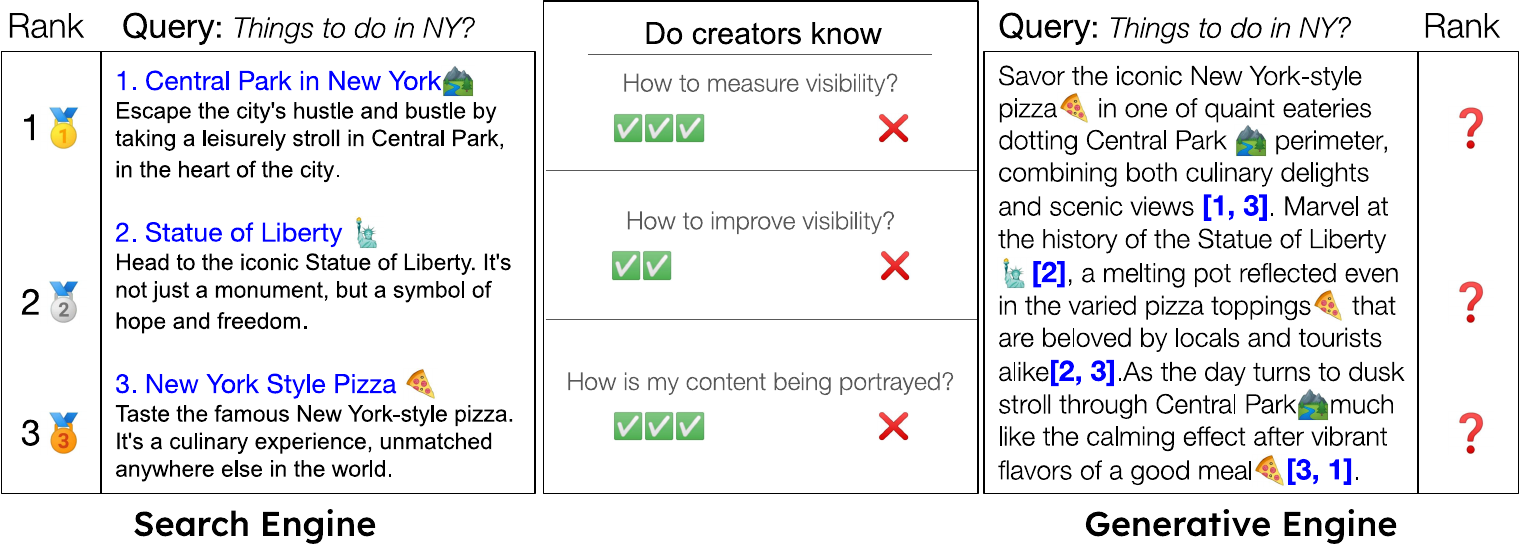}
    \caption{Ranking and Visibility Metrics are straightforward in traditional search engines, which list website sources in ranked order with verbatim content. However, \Le{}s generate rich, structured responses, often embedding citations in a single block interleaved with each other. This makes ranking and visibility nuanced and multi-faceted. Further, unlike search engines, where significant research has been conducted on improving visibility, optimizing visibility in \lee{} responses remains unclear. To address these challenges, our black-box optimization framework proposes a series of well-designed impression metrics that creators can use to \textit{gauge} and \textit{optimize} their website's performance and also allows the creator to define their impression metrics.}
    \label{fig:geo_explain}
\end{figure*}

Despite the deployment of numerous \lee{}s to millions of users, there is currently no standard framework. We provide a formulation that accommodates various modular components in their design.
We describe a \lee{}, which includes several backend generative models and a search engine for source retrieval. A \Le{} (\LE{}) takes a user query $q_{u}$ and returns a natural language response $r$, where $P_{U}$ represents personalized user information. The \LE{} can be represented as a function:
\begin{equation}
\label{eq:le_single_turn}
 f_{GE} := (q_{u}, P_{U}) \rightarrow r
\end{equation}

\Le{}s comprise two crucial components: a.) A set of generative models $G = \{G_1, G_2 ... G_n\}$, each serving a specific purpose like query reformulation or summarization, and b.) A search engine $SE$ that returns a set of sources $S = \{s_1, s_2 ... s_m\}$ given a query $q$. 
We present a representative workflow in Figure~\ref{fig:le_overview}, which, at the time of writing, closely resembles the design of BingChat.
This workflow breaks down the input query into a set of simpler queries that are easier to consume for the search engine. 
Given a query, a query re-formulating generative model, $G_1 = G_{qr}$, generates a set of queries $Q^1 = \{q_1, q_2 ... q_n\}$, which are then passed to the search engine $SE$ to retrieve a set of ranked sources $S = \{s_1, s_2, ..., s_m\}$.
The sets of sources $S$ are passed to a summarizing model $G_2 = G_{sum}$, which generates a summary $Sum_j$ for each source in $S$, resulting in the summary set ($Sum = \{Sum_1, Sum_2, ..., Sum_m\}$).
The summary set is passed to a response-generating model $G_3 = G_{resp}$, which generates a cumulative response $r$ backed by sources $S$. 
In this work, we focus on single-turn \Le{}s, but the formulation can be extended to multi-turn Conversational \Le{}s (Appendix~\ref{app:conversationaL_le}).

The response $r$ is typically a structured text with embedded citations. Citations are important given the tendency of LLMs to hallucinate information~\cite{ji2023survey}. Specifically, consider a response $r$ composed of sentences $\{l_1, l_2 ... l_o\}$. Each sentence may be backed by a set of citations that are part of the retrieved set of documents $C_i \subset S$. An ideal \lee{} should ensure all statements in the response are supported by relevant citations (high citation recall), and all citations accurately support the statements they're associated with (high citation precision)~\cite{Liu2023EvaluatingVI}. We refer readers to Figure~\ref{fig:geo_explain} for a representative \lee{} response.

\subsection{\leo}

The advent of search engines led to search engine optimization (SEO), a process to help website creators optimize their content to improve search engine rankings. 
Higher rankings correlate with increased visibility and website traffic. However, traditional SEO methods are not directly applicable to \Le{}s. This is because, unlike traditional search engines, the generative model in \lee{}s is not limited to keyword matching, and the use of language models in ingesting source documents and response generation results in a more nuanced understanding of text documents and user query. With \lee{}s rapidly emerging as the primary information delivery paradigm 
and SEO is not directly applicable; new techniques are needed.
To this end, we propose \leo{}, a new paradigm where content creators aim to increase their visibility (or impression) in \lee{} responses.
We define the visibility of a website (also referred to as a citation) $c_i$ in a cited response $r$ by the function $Imp(c_i, r)$, which the website creator wants to maximize.
From the \lee{}'s perspective, the goal is to maximize the visibility of citations most relevant to the user query, i.e., maximize $\sum_{i} f (Imp(c_i, r), Rel(c_i, q, r)$), where $Rel(c_i, q, r)$ measures the relevance of citation $c_i$ to the query $q$ in the context of response $r$ and $f$ is determined by the exact algorithmic design of \lee{} and is a black-box function to end-users.
Further, both the functions $Imp$ and $Rel$ are subjective and not well-defined yet for \lee{}s, and we define them next.

\subsubsection{Impressions for \Le{}s}
\label{sec:impression}

In SEO, a website's impression (or visibility) is determined by its average ranking over a range of queries. However, \lee{}s' output nature necessitates different impression metrics. Unlike search engines, \Le{}s combine information from multiple sources in a single response. Factors such as length, uniqueness, and presentation of the cited website determine the true visibility of a citation. Thus, as illustrated in Figure~\ref{fig:geo_explain}, while a simple ranking on the response page serves as an effective metric for impression and visibility in conventional search engines, such metrics are not applicable to \lee{} responses.

In response to this challenge, we propose a suite of impression metrics designed with three key principles in mind: 1.) The metrics should hold relevance for creators, 2.) They should be explainable, and 3.) They should be easily comprehensible by a broad spectrum of content creators. 
The first of these metrics, the ``Word Count'' metric, is the normalized word count of sentences related to a citation. Mathematically, this is defined as:
\begin{equation}
 Imp_{wc}(c_i, r) = \frac{\sum_{s \in S_{c_i}} |s|}{\sum_{s \in S_{r}} |s|}
\end{equation}
Here $S_{c_i}$ is the set of sentences citing $c_i$, $S_{r}$ is the set of sentences in the response, and $|s|$ is the number of words in sentence $s$. In cases where a sentence is cited by multiple sources, we share the word count equally with all the citations. Intuitively, a higher word count correlates with the source playing a more important part in the answer, and thus, the user gets higher exposure to that source. However, since ``Word Count'' is not impacted by the ranking of the citations (whether it appears first, for example), we propose a position-adjusted count that reduces the weight by an exponentially decaying function of the citation position:
\begin{equation}
 Imp_{pwc}(c_i, r) = \frac{\sum_{s \in S_{c_i}} |s| \cdot e^{-\frac{pos(s)}{|S|}}}{\sum_{s \in S_{r}} |s|}
\end{equation}
Intuitively, sentences that appear first in the response are more likely to be read, and the exponent term in definition $Imp_{pwc}$ gives higher weightage to such citations. Thus, a website cited at the top may have a higher impression despite having a lower word count than a website cited in the middle or end of the response. Further, the choice of exponentially decaying function is motivated by several studies showing click-through rates follow a power-law as a function of ranking in search engines~\cite{searchenginewatch2011,dean2023}. 
While the above impression metrics are objective and well-grounded, they ignore the subjective aspects of the impact of citations on the user's attention. To address this, we propose the "Subjective Impression" metric, which incorporates facets such as the relevance of the cited material to the user query, influence of the citation, uniqueness of the material presented by a citation, subjective position, subjective count, probability of clicking the citation, and diversity in the material presented. We use G-Eval~\cite{Liu2023GEvalNE}, the current state-of-the-art for evaluation with LLMs, to measure each of these sub-metrics.

\subsubsection{\leo{} methods for website}
\label{sec:leo_methods}

To improve impression metrics, content creators must make changes to their website content. 
We present several \lee{}-agnostic strategies, referred to as \leo{} methods (\LEO{}). Mathematically, every \LEO{} method is a function $f: W \rightarrow W'_i$, where $W$ is the initial web content, and $W'$ is the modified content after applying the \LEO{} method. The modifications can range from simple stylistic alterations to incorporating new content in a structured format. A well-designed \LEO{} is equivalent to a black-box optimization method that, without knowing the exact algorithmic design of \lee{}s, can increase the website's visibility and implement textual modifications to $W$ independent of the exact queries. 

For our experiments, we apply \leo{} methods on website content using a large language model, prompted to perform specific stylistic and content changes to the website. In particular, based on the \LEO{} method defining a specific set of desired characteristics, the source content is modified accordingly. We propose and evaluate several such methods:

\textbf{1: \leoauthoritative{}:} Modifies text style of the source content to be more persuasive and authoritative, \textbf{2. \leostats{}:} Modifies content to include quantitative statistics instead of qualitative discussion, wherever possible, \textbf{3. \leoseo{}:} Modifies content to include more keywords from the query, as expected in classical SEO optimization. 
\textbf{4. \leociting{} \& 5. \leoquotes{}:} Adds relevant citations and quotations from credible sources respectively, 6.) \textbf{6. \leosimple{}:} Simplifies the language of website, while \textbf{7. \leofluent{}} improves the fluency of website text. \textbf{8. \leounique{} \& 9. \leotechnical:} involves adding unique and technical terms respectively wherever possible, 

These methods cover diverse general strategies that website owners can implement quickly and use regardless of the website content. Further, except for methods 3, 4, and 5, the remaining methods enhance the presentation of existing content to increase its persuasiveness or appeal to the \lee{}, without requiring extra content. On the other hand, methods 3,4 and 5 may require some form of additional content. To analyze the performance gain of our methods, for each input user query, we randomly select one source website to be optimized and apply each of the \LEO{} methods separately on the same source. We refer readers to Appendix~\ref{app:meth} for more details on \LEO{} methods.
\section{Experimental Setup}

\subsection{Evaluated \Le{}}

In accordance with previous works~\cite{Liu2023EvaluatingVI}, we use a 2-step setup for \Le{} design.
The first step involves fetching relevant sources for input query, followed by a second step where an \llm{} generates a response based on the fetched sources. Similar to prevous works, we do not use summarization and provide the whole response for each source. Due to context length limitations and quadratic scaling cost based on the context size of transformer models, only the top 5 sources are fetched from the Google search engine for every query. The setup closely mimics the workflow used in previous works and the general design adopted by commercial \LE{}s such as you.com and perplexity.ai.
The answer is then generated by the gpt3.5-turbo model~\cite{gpt35turbo} using the same prompt as prior work~\cite{Liu2023EvaluatingVI}. We sample 5 different responses at temperature=0.7, to reduce statistical deviations.

Further in Section~\ref{app:perplexity_results}, we evaluate the same \leo{} methods on Perplexity.ai, which is a commercially deployed generative engine, highlighting the generalizability of our proposed \leo{} methods.

\subsection{Benchmark : \bench{}}
Since there is currently no publicly available dataset containing \Le{} related queries, we curate \textbf{\bench}, a benchmark consisting of 10K queries from multiple sources, repurposed for \lee{}s, along with synthetically generated queries. The benchmark includes queries from nine different sources, each further categorized based on their target domain, difficulty, query intent, and other dimensions. 

\paragraph{Datasets:}
\textbf{1. MS Macro, 2. ORCAS-1, and 3. Natural Questions:}~\cite{Kwiatkowski2019NaturalQA, Alexander2022ORCASIQA, Craswell2021MSMB} These datasets contain real anonymized user queries from Bing and Google Search Engines. 
These three collectively represent the common set of datasets that are used in search engine related research. However, \Le{}s will be posed with far more difficult and specific queries with the intent of synthesizing answers from multiple sources instead of searching for them. To this end, we repurpose several other publicly available datasets:
\textbf{4. AllSouls:} This dataset contains essay questions from "All Souls College, Oxford University." The queries in this dataset require \Le{}s to perform appropriate reasoning to aggregate information from multiple sources.
\textbf{5. LIMA:}~\cite{Zhou2023LIMALI} contains challenging questions requiring \Le{}s to not only aggregate information but also perform suitable reasoning to answer the question (e.g., writing a short poem, python code.). 
\textbf{6. Davinci-Debtate}~\cite{Liu2023EvaluatingVI} contains debate questions generated for testing \Le{}s.
\textbf{7. Perplexity.ai Discover\footnote{\url{https://www.perplexity.ai/discover}}:} These queries are sourced from Perplexity.ai's Discover section, which is an updated list of trending queries on the platform.
\textbf{8. ELI-5\footnote{\url{https://huggingface.co/datasets/eli5_category}}:} This dataset contains questions from the ELI5 subreddit, where users ask complex questions and expect answers in simple, layman's terms.
\textbf{9. GPT-4 Generated Queries:} To supplement diversity in query distribution, we prompt GPT-4~\cite{openai2024gpt4} to generate queries ranging from various domains (e.g., science, history) and based on query intent (e.g., navigational, transactional) and based on difficulty and scope of generated response (e.g., open-ended, fact-based).

\paragraph{}Our benchmark comprises 10K queries divided into 8K, 1K, and 1K for train, validation, and test splits, respectively. We preserve the real-world query distribution, with our benchmark containing 80\% informational queries and 10\% each for transactional and navigational queries. Each query is augmented with the cleaned text content of the top 5 search results from the Google search engine.

\paragraph{Tags} Optimizing website content often requires targeted changes based on the task's domain. Additionally, a user of \leo{} may need to identify an appropriate method for only a subset of queries, considering multiple factors such as domain, user intent, and query nature. To facilitate this, we tag each query with one of seven different categories. For tagging, we employ the GPT-4 model and manually verify high recall and precision on the test split.

Overall, \bench{} consists of queries from 25 diverse domains such as Arts, Health, and Games; it features a range of query difficulties from simple to multi-faceted; includes 9 different types of queries such as informational and transactional; and encompasses 7 different categorizations. Owing to its specially designed high diversity, the size of the benchmark, and its real-world nature, \bench{} is a comprehensive benchmark for evaluating \Le{}s and serves as a standard testbed for assessing them for various purposes in this and future works. We provide more details about \bench{} in Appendix~\ref{app:exp_setup_bench}.

\begin{table*}
\centering
\resizebox{0.99\linewidth}{!}{%
\begin{tabular}{lccccccccccc}
\toprule
\multirow{2}{*}[-6pt]{\large{\bf{Method}}} & \multicolumn{3}{c}{\bf{\wordposmetric}}  & \multicolumn{8}{c}{\bf{\subjectiveimpression}} \\
\cmidrule(lr){2-4} \cmidrule(lr){5-12}
& Word & Position & \bf{Overall} & Rel. & Infl. & Unique  & Div. & FollowUp & Pos. & Count & \bf{Average}
\\
\midrule
\rowcolor{gray!20} \multicolumn{12}{c}{Performance without \leo{}} \\ \midrule
\bf{\leobaseline{}} & $19.5$ & $19.3$ & $19.3$ & $19.3$ & $19.3$ & $19.3$ & $19.3$ & $19.3$ & $19.3$ & $19.3$ & $19.3$ \\
\midrule
\rowcolor{gray!20} \multicolumn{12}{c}{Non-Performing \leo{} methods} \\ \midrule
\bf{\leoseo{}} & $17.8$ & $17.7$ & $17.7$ & $19.8$ & $19.1$ & $20.5$ & $20.4$ & $20.3$ & $20.5$ & $20.4$ & $20.2$ \\
\bf{\leounique{}} & $20.7$ & $20.5$ & $20.5$ & $20.5$ & $20.1$ & $19.9$ & $20.4$ & $20.2$ & $20.7$ & $20.2$ & $20.4$ \\

\midrule
\rowcolor{gray!20} \multicolumn{12}{c}{High-Performing \leo{} methods} \\ \midrule
\bf{\leosimple{}} & $22.2$ & $22.4$ & $22.0$ & $20.2$ & $21.0$ & $20.0$ & $20.1$ & $20.1$ & $20.9$ & $19.9$ & $20.5$ \\
\bf{\leoauthoritative{}} & $21.8$ & $21.3$ & $21.3$ & $22.3$ & $22.1$ & $22.4$ & $23.1$ & $22.2$ & $23.1$ & $22.7$ & $22.9$ \\
\bf{\leotechnical{}} & $23.1$ & $22.7$ & $22.7$ & $20.9$ & $21.7$ & $20.5$ & $21.2$ & $20.8$ & $21.9$ & $20.8$ & $21.4$ \\
\bf{\leofluent{}} & $25.1$ & $24.6$ & $24.7$ & $21.1$ & $22.9$ & $20.4$ & $21.6$ & $21.0$ & $22.4$ & $21.1$ & $21.9$ \\
\bf{\leociting{}} & $24.9$ & $24.5$ & $24.6$ & $21.4$ & $22.5$ & $21.0$ & $21.6$ & $21.2$ & $22.2$ & $20.7$ & $21.9$ \\
\bf{\leoquotes{}} & $27.8$ & $27.3$ & $\textbf{27.2}$ & $23.8$ & $25.4$ & $23.9$ & $24.4$ & $22.9$ & $24.9$ & $23.2$ & $\textbf{24.7}$ \\
\bf{\leostats{}} & $25.9$ & $25.4$ & $25.2$ & $22.5$ & $24.5$ & $23.0$ & $23.3$ & $21.6$ & $24.2$ & $23.0$ & $23.7$ \\
\bottomrule
\end{tabular}
}
\caption{\label{results-main}
Absolute impression metrics of \LEO{} methods on \bench{}. Performance Measured on Two metrics and their sub-metrics. Compared to baselines, simple methods like \leoseo{} traditionally used in SEO don't perform well. However, our proposed methods such as \leostats{} and \leoquotes{} show strong performance improvements across all metrics. The best methods improve upon baseline by 41\% and 28\% on \wordposmetric{} and \subjectiveimpression{} respectively. For readability, \subjectiveimpression{} scores are normalized with respect to \wordposmetric{} resulting in similar baseline scores.}
\label{tab:results-main}
\end{table*}

\subsection{\LEO{} Methods}

We evaluate 9 different proposed \LEO{} methods as described in Section~\ref{sec:leo_methods}. We compare them with a baseline, which measures the impression metric of unmodified website sources. We evaluate methods on the complete \bench{} test split. Further, to reduce variance in results, we run our experiments on five different random seeds and report the average. 

\subsection{Evaluation Metrics}
We utilize the impression metrics as defined in Section~\ref{sec:impression}. Specifically, we employ two impression metrics: \textbf{1. Position-Adjusted Word Count}, which combines word count and position count. To analyze the effect of individual components, we also report scores on the two sub-metrics separately. \textbf{2. Subjective Impression}, which is a subjective metric encompassing seven different aspects: 1) relevance of the cited sentence to the user query, 2) influence of the citation, assessing the extent to which the generated response relies on the citation, 3) uniqueness of the material presented by a citation, 4) subjective position, gauging the prominence of the positioning of source from the user's viewpoint, 5) subjective count, measuring the amount of content presented from the citation as perceived by the user, 6) likelihood of the user clicking the citation, and 7) diversity of the material presented. These sub-metrics assess diverse aspects that content creators can target to improve one or more areas effectively. Each sub-metric is evaluated using GPT-3.5, following a methodology akin to that described in G-Eval~\cite{Liu2023GEvalNE}. In G-Eval, a form-based evaluation template is provided to the language model, along with a \LE{} generated response with citations. The model outputs a score (computed by sampling multiple times) for each citation. However, since G-Eval scores are poorly calibrated, we normalize them to have the same mean and variance as \wordposmetric{} to enable a fair and meaningful comparison. We provide the exact templates used in Appendix~\ref{app:impression}.

Furthermore, all impression metrics are normalized by multiplying them with a constant factor so that the sum of the impressions of all citations in a response equals 1. In our analysis, we compare methods by calculating the relative improvement in impression. For an initial generated response $r$ from sources $S_i\in\{s_1,\dots,s_m\}$, and a modified response $r'$, the relative improvement in impression for each source $s_i$ is measured as:
\begin{equation}
 Improvement_{s_i} = \frac{Imp_{s_i}(r') - Imp_{s_i}(r)}{Imp_{s_i}(r)} \times 100
\end{equation}
The modified response $r'$ is produced by applying the \LEO{} method being evaluated to one of the sources $s_i$. The source $s_i$ selected for optimization is chosen randomly but remains constant for a particular query across all \LEO{} methods.

\section{Results}
\label{sec:main_results}

We evaluate various \leo{} methods designed to optimize website content for better visibility in \Le{} responses, compared against a baseline with no optimization. Our evaluation used \bench{}, a diverse benchmark of user queries from multiple domains and settings. Performance was measured using two metrics: \textit{\wordposmetric{}} and \textit{\subjectiveimpression{}}. The former considers word count and citation position in the \LE{}'s response, while the latter computes multiple subjective factors, giving an overall impression score.

Table~\ref{tab:results-main} details the absolute impression metrics of different methods on multiple metrics. The results reveal that our \LEO{} methods consistently outperform the baseline across all metrics on \bench{}. This shows the robustness of these methods to varying queries, yielding significant improvements despite query diversity. Specifically, our top-performing methods, \leociting{}, \leoquotes{}, and \leostats{}, achieved a relative improvement of 30-40\% on the \textit{\wordposmetric{}} metric and 15-30\% on the \textit{\subjectiveimpression{}} metric.
These methods, involving adding relevant statistics (\leostats{}), incorporating credible quotes (\leoquotes{}), and including citations from reliable sources (\leociting{}) in the website content, require minimal changes but significantly improve visibility in \LE{} responses, enhancing both the credibility and richness of the content.

\begin{table*}[!t]
  \centering
  \begin{minipage}{.48\linewidth}
    \centering
    \resizebox{\linewidth}{!}{%
    \begin{tabular}{lccccc}
      \toprule
      \multirow{2}{*}[-6pt]{\large{\bf{Method}}} & \multicolumn{5}{c}{\bf{Relative Improvement (\%) in Visibility}} \\
      \cmidrule(lr){2-6} 
      & Rank-1 & Rank-2 & Rank-3 & Rank-4 & Rank-5 \\
      \midrule
      \textbf{\leoauthoritative{}} & -6.0 & 4.1 & -0.6 & 12.6 & 6.1 \\
      \textbf{Fluency  Opt.} & -2.0 & 5.2 & 3.6 & -4.4 & 2.2 \\
      \textbf{\leociting{}} & -30.3 & 2.5 & 20.4 & 15.5 & 115.1 \\
      \textbf{\leoquotes{}} & -22.9 & -7.0 & 3.5 & 25.1 & 99.7  \\
      \textbf{\leostats{}} & -20.6 & -3.9 & 8.1 & 10.0 & 97.9 \\
      \bottomrule
    \end{tabular}
    }
    \caption{Visibility changes through \LEO{} methods for sources with different Rankings in Search Engine. \LEO{} is especially helpful for lower ranked websites.}
    \label{tab:analysis-allopt}
  \end{minipage}\hfill
  \begin{minipage}{.48\linewidth}
    \centering
    \resizebox{\linewidth}{!}{%
    \begin{tabular}{lccc}
      \toprule
      \multirow{2}{*}[-6pt]{\large{\bf{Method}}} & \multicolumn{3}{c}{\bf{Top Performing Tags}} \\
      \cmidrule(lr){2-4} 
      & Rank-1 & Rank-2 & Rank-3 \\
      \midrule
      \textbf{\leoauthoritative{}} & Debate & History & Science \\
      \textbf{Fluency  Opt.} & Business & Science & Health \\
      \textbf{\leociting{}} & Statement & Facts & Law \& Gov. \\
      \textbf{\leoquotes{}} & People \& Society & Explanation & History \\
      \textbf{\leostats{}} & Law \& Gov. & Debate & Opinion \\
      \bottomrule
    \end{tabular}
    }
    \caption{Top Performing categories for each of the \LEO{} methods. Website-owners can choose relevant \LEO{} strategy based on their target domain.}
    \label{tab:analysis-tags}
  \end{minipage}
\end{table*}

Interestingly, stylistic changes such as improving fluency and readability of the source text (\leofluent{} and \leosimple{}) also resulted in a significant visibility boost of 15-30\%. This suggests that \Le{}s value not only content but also information presentation.

Further, given generative models are often designed to follow instructions, one would expect a more persuasive and authoritative tone in website content to boost visibility. 
However, we find no significant improvement, demonstrating that \Le{}s are already somewhat robust to such changes. This highlights the need for website owners to focus on improving content presentation and credibility.

Finally, we evaluate keyword stuffing, i.e., adding more relevant keywords to website content. While widely used for Search Engine Optimization, we find such methods offer little to no improvement on \lee{}'s responses. This underscores the need for website owners to rethink optimization strategies for \lee{}s, as techniques effective in search engines may not translate to success in this new paradigm.
\section{Analysis}

\subsection{Domain-Specific \leo{}s}

In Section~\ref{sec:main_results}, we presented the improvements achieved by \LEO{} across the entirety of the \bench{} benchmark. However, in real-world SEO scenarios, domain-specific optimizations are often applied. With this in mind, and considering that we provide categories for every query in \bench{}, we delve deeper into the performance of various \LEO{} methods across these categories.

Table~\ref{tab:analysis-tags} provides a detailed breakdown of the categories where our \LEO{} methods have proven to be most effective. A careful analysis of these results reveals several intriguing observations. 
For instance, \leoauthoritative{} significantly improves performance in debate-style questions and queries related to the ``historical'' domain. This aligns with our intuition, as a more persuasive form of writing is likely to hold more value in debates.

Similarly, the addition of citations through \leociting{} is particularly beneficial for factual questions, likely because citations provide a source of verification for the facts presented, thereby enhancing the credibility of the response.
The effectiveness of different \LEO{} methods varies across domains. For example, as shown in row 5 of Table~\ref{tab:analysis-tags}, domains such as `Law \& Government' and question types like `Opinion' benefit significantly from the addition of relevant statistics in the website content, as implemented by \leostats{}. This suggests that data-driven evidence can enhance the visibility of a website in particular contexts.
The method \leoquotes{} is most effective in the `People \& Society,' `Explanation,' and `History' domains. This could be because these domains often involve personal narratives or historical events, where direct quotes can add authenticity and depth to the content.
Overall, our analysis suggests that website owners should strive towards making domain-specific targeted adjustments to their websites for higher visibility.

\begin{figure}[t]
    \centering
    \includegraphics[width=0.99\linewidth]{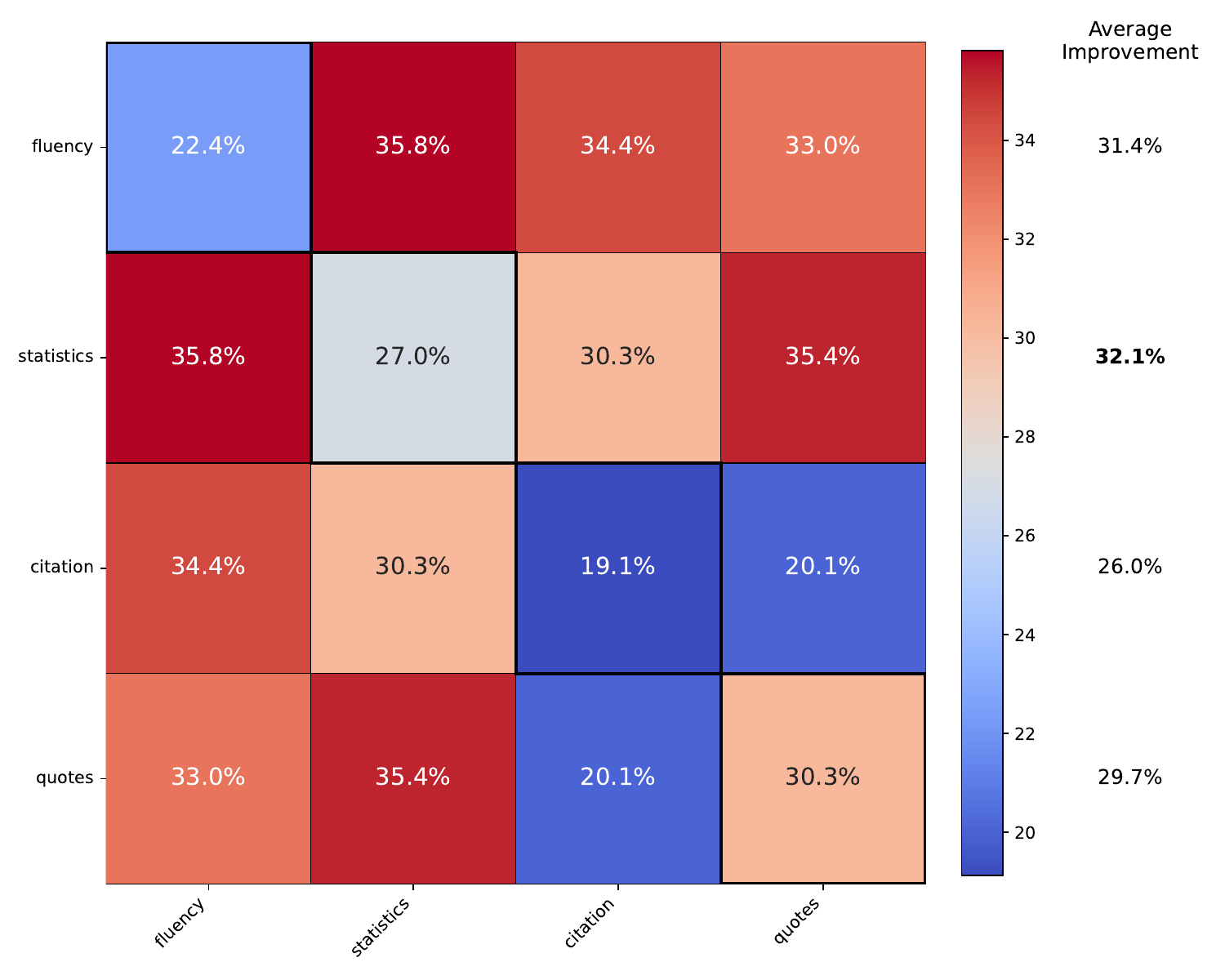}
    \caption{Relative Improvement on using combination of \LEO{} strategies. Using \leofluent{} and \leostats{} in conjunction results in maximum performance. The rightmost column shows using \leofluent{} with other strategies is most beneficial.}
    \label{fig:comb}
\end{figure}

\begin{table*}[!t]
\centering
\resizebox{\linewidth}{!}{%
\begin{tabular}{llc}
\toprule
\multirow{1}{*}{\bf{Method}} & \multicolumn{1}{c}{\bf{\LEO{} Optimization}} & \bf{Relative Improvement} \\
\midrule

& \textbf{Query:} What is the secret of Swiss chocolate  & \\
& \multirow{4}{0.75\linewidth}{With per capita annual consumption averaging between 11 and 12 kilos, Swiss people rank among the top chocolate lovers in the world \colorbox{greenback}{(According to a survey conducted by The International Chocolate}  \colorbox{greenback}{Consumption Research Group \textcolor{blue}{[1]})}} & \\ 
\bf{\leociting{}} & & \bf{132.4\%} \\
& &  \\
& &  \\
\midrule
& \textbf{Query:} Should robots replace humans in the workforce? & \\
& \multirow{4}{0.75\linewidth}{\textbf{Source:} Not here, and not now — until recently. The big difference is that the robots have come not to destroy our lives, but to disrupt our work, \colorbox{greenback}{with a staggering 70\% increase in robotic involvement in the last decade}.} & \\ 
\bf{\leostats{}} & & \bf{65.5\%}  \\
& & \\
& & \\

\midrule
& \textbf{Query:} Did the jacksonville jaguars ever make it to the superbowl? & \\
& \multirow{4}{0.75\linewidth}{\textbf{Source:} \colorbox{greenback}{It is important to note that} The Jaguars have never \colorbox{redback}{appeared} \colorbox{greenback}{made an appearance} in the Super Bowl. \colorbox{greenback}{However,} They have \colorbox{greenback}{achieved} \colorbox{greenback}{an impressive feat by securing} 4 divisional titles \colorbox{redback}{to their name.}  \colorbox{greenback}{, a} \colorbox{greenback}{testament to their prowess and determination.}} & \\ 
\bf{\leoauthoritative{}} & & \bf{89.1\%} \\
& &  \\
& &  \\

\bottomrule

\end{tabular}
} 
\caption{
\label{table:qualitative}
Representative examples of \LEO{} methods optimizing source website. \colorbox{greenback}{Additions} are marked in green and \colorbox{redback}{Deletions} in red. Without adding any substantial new information, \LEO{} methods significantly increase the visibility of the source content.}
\end{table*}

\subsection{Optimization of Multiple Websites}

In the evolving landscape of \Le{}s, \LEO{} methods are expected to become widely adopted, leading to a scenario where all source contents are optimized using \LEO{}. To understand the implications, we conducted an evaluation of \LEO{} methods by optimizing all source contents simultaneously, with results presented in Table~\ref{tab:analysis-allopt}. A key observation is the differential impact of \LEO{} on websites based on their Search Engine Results Pages (SERP) ranking. Notably, lower-ranked websites, which typically struggle for visibility, benefit significantly more from \LEO{}. This is because traditional search engines rely on multiple factors, such as the number of backlinks and domain presence, which are challenging for small creators to achieve. However, since \Le{}s utilize generative models conditioned on website content, factors such as backlink building should not disadvantage small creators. This is evident from the relative improvements in visibility shown in Table~\ref{tab:analysis-allopt}. For example, the \leociting{} method led to a substantial 115.1\% increase in visibility for websites ranked fifth in SERP, while on average, the visibility of the top-ranked website decreased by 30.3\%.

This finding highlights \LEO{}'s potential as a tool to democratize the digital space. Many lower-ranked websites are created by small content creators or independent businesses, who traditionally struggle to compete with larger corporations in top search engine results. The advent of \Le{}s might initially seem disadvantageous to these smaller entities. However, the application of \LEO{} methods presents an opportunity for these content creators to significantly improve their visibility in \Le{} responses. By enhancing their content with \LEO{}, they can reach a wider audience, leveling the playing field and allowing them to compete more effectively with larger corporations.


\subsection{Combination of \LEO{} Strategies}

While individual \LEO{} strategies show significant improvements across various domains, in practice, website owners are expected to employ multiple strategies in conjunction. To study the performance improvements achieved by combining \LEO{} strategies, we consider all pairs of combinations of the top 4 performing \LEO{} methods, namely \leociting{}, \leofluent{}, \leostats{}, and \leoquotes{}. Figure~\ref{fig:comb} displays the heatmap of relative improvement in the \wordposmetric{} visibility metric achieved by combining different \LEO{} strategies. The analysis demonstrates that the combination of \leo{} methods can enhance performance, with the best combination (\leofluent{} and \leostats{}) outperforming any single \LEO{} strategy by more than 5.5\%\footnote{Due to cost constraints, the analysis was conducted on a subset of 200 examples from the test split, and therefore the numbers presented here differ from those in Table~\ref{tab:results-main}}. Furthermore, \leociting{} significantly boosts performance when used in conjunction with other methods (Average: 31.4\%), despite it being relatively less effective when used alone (8\% lower than \leoquotes{}). The findings underscore the importance of studying \LEO{} methods in combination, as they are likely 
to be used by content creators in the real world.

\subsection{Qualitative Analysis}

We present a qualitative analysis of \LEO{} methods in Table~\ref{table:qualitative}, containing representative examples where \LEO{} methods boost source visibility with minimal changes. 
Each method optimizes a source through suitable text additions and deletions. In the first example, we see that simply adding the source of a statement can significantly boost visibility in the final answer, requiring minimal effort from the content creator. The second example demonstrates that adding relevant statistics wherever possible ensures increased source visibility in the final \Le{} response. Finally, the third row suggests that merely emphasizing parts of the text and using a persuasive text style can also lead to improvements in visibility.

\section{\LEO{} in the Wild : Experiments with Deployed \Le{}}

\begin{table}[!htbp]
\centering
\resizebox{\linewidth}{!}{%
\begin{tabular}{lcc}
\toprule
\multirow{1}{*}[-6pt]{\large{\bf{Method}}} & \multicolumn{1}{c}{\bf{\wordposmetric}}  & \multicolumn{1}{c}{\bf{\subjectiveimpression}} \\
\cmidrule(lr){2-3} 
\midrule
\bf{\leobaseline{}} & $24.1$ & $24.7$ \\
\midrule
\bf{\leoseo{}} & $21.9$ & $28.1$ \\
\midrule
\bf{\leoquotes{}} & $\textbf{29.1}$ & $32.1$ \\
\bf{\leostats{}} & $26.2$ & $\textbf{33.9}$ \\

\bottomrule
\end{tabular}
}
\caption{\label{results-main-perplexity-small}
Absolute impression metrics of \LEO{} methods on \bench{} with Perplexity.ai as \LE{}. While SEO methods such as \leoseo{} perform poorly, our proposed \LEO{} methods generalize well to multiple \lee{}s significanlty improve content visibility. \vspace{-4mm}}
\label{tab:results-perplexity-small}
\end{table}

To reinforce the efficacy of our proposed \leo{} methods, we evaluate them on Perplexity.ai, a real deployed \Le{} with a large user base. Results are in Table~\ref{tab:results-perplexity-small}. Similar to our \lee{}, \leoquotes{} performs best in \wordposmetric{} with a 22\% improvement over the baseline. Methods that performed well in our \lee{} such as \leociting{}, \leostats{} show improvements of up to 9\% and 37\% on the two metrics. Our observations, such as the ineffectiveness of traditional SEO methods like \leoseo{}, are further highlighted, as it performs 10\% worse than the baseline. The results are significant for three reasons: 1) they underscore the importance of developing different \leo{} methods to benefit content creators, 2) they highlight the generalizability of our proposed \LEO{} methods on different \lee{}s, 3) they demonstrate that content creators can use our easy-to-implement proposed \LEO{} methods directly, thus having a high real-world impact. We refer readers to Appendix~\ref{app:perplexity_results} for more details.

\section{Related Work}
\label{sec:related_work}

\paragraph{Evidence-based Answer Generation:} Previous works have used several techniques for answer generation backed by sources. \citet{Nakano2021WebGPTBQ} trained GPT-3 to navigate web environments to generate source-backed answers. Similarly, other methods~\cite{Shuster2022BlenderBot3A, Thoppilan2022LaMDALM, Menick2022TeachingLM} fetch sources via search engines for answer generation. Our work unifies these approaches and provides a common benchmark for improving these systems in the future. In a recent working draft, Kumar and Lakkaraju~\cite{kumar2024manipulating} showed that strategic text sequences can manipulate LLM recommendations to enhance product visibility in \lee{}s. While their approach focuses on increasing product visibility through adversarial text, our method introduces non-adversarial strategies to optimize any website content for improved visibility in \lee{} search results.

\paragraph{Retrieval-Augmented Language Models:} Several recent works have tackled the issues of limited memory of language models by fetching relevant sources from a knowledge base to complete a task~\cite{Asai2021OneQA, Mialon2023AugmentedLM, Guu2020REALMRL}.
However, \Le{} needs to generate an answer and provide attributions throughout the answer. Further, \Le{} is not limited to a single text modality regarding both input and output. Additionally, the framework of \Le{} is not limited to fetching relevant sources but instead comprises multiple tasks such as query reformulation, source selection, and making decisions on how and when to perform them.

\paragraph{Search Engine Optimization:} In nearly the past 25 years, extensive research has optimized web content for search engines~\cite{Ankalkoti2017SurveyOS, Shahzad2020TheNT, Kumar2019ASO}. These methods fall into On-Page SEO, improving content and user experience, and Off-Page SEO, boosting website authority through link building. In contrast, \LEO{} deals with a more complex environment involving multi-modality, conversational settings. Since \LEO{} is optimized against a generative model not limited to simple keyword matching, traditional SEO strategies will not apply to \Le{} settings, highlighting the need for \LEO{}. 
\section{Conclusion}
\label{sec:conclusion}

In this work, we formulate search engines augmented with generative models that we dub \lee{}s. 
We propose \leo{} (\LEO{}) to empower content creators to optimize their content under \lee{}s.
We define impression metrics for \lee{}s and propose and release \bench{}: a benchmark encompassing diverse user queries from multiple domains and settings, along with relevant sources needed to answer those queries.
We propose several ways to optimize content for \lee{}s and demonstrate that these methods can boost source visibility by up to 40\% in \lee{} responses.
Among other findings, we show that including citations, quotations from relevant sources, and statistics can significantly boost source visibility.
Further, we discover a dependence of \LEO{} methods' effectiveness on the query domain and the potential of combining multiple \LEO{} strategies in conjunction. We show promising results on a commercially deployed \lee{} with millions of active users, showcasing the real-world impact of our work. 
In summary, our work is the first to formalize the important and timely \LEO{} paradigm, releasing algorithms and infrastructure (benchmarks, datasets, and metrics) to facilitate rapid progress in generative engines by the community. This serves as a first step towards understanding the impact of \lee{}s on the digital space and the role of \LEO{} in this new paradigm of search engines.

\section{Limitations}

While we rigorously test our proposed methods on two generative engines, including a publicly available one, methods may need to adapt over time as \LE{}s evolve, mirroring the evolution of SEO. Additionally, despite our efforts to ensure the queries in our \bench{} closely resemble real-world queries, the nature of queries can change over time, necessitating continuous updates. Further, owing to the black-box nature of search engine algorithms, we didn't evaluate how \LEO{} methods affect search rankings. However, we note that changes made by \LEO{} methods are targeted changes in textual content, bearing some resemblance with SEO methods, while not affecting other metadata such as domain name, backlinks, etc, and thus, they are less likely to affect search engine rankings. Further, as larger context lengths in language models become economical, it is expected that future generative models will be able to ingest more sources, thus reducing the impact of search rankings. 
Lastly, while every query in our proposed \bench{} is tagged and manually inspected, there may be discrepancies due to subjective interpretations or errors in labeling. 



\section{Acknowledgements}

This material is based upon work supported by the National Science Foundation under Grant No. 2107048. Any opinions, findings, and conclusions or recommendations expressed in this material are those of the author(s) and do not necessarily reflect the views of the National Science Foundation. 

\bibliographystyle{ACM-Reference-Format}
\bibliography{custom}


\begin{thebibliography}{25}


\ifx \showCODEN    \undefined \def \showCODEN     #1{\unskip}     \fi
\ifx \showDOI      \undefined \def \showDOI       #1{#1}\fi
\ifx \showISBNx    \undefined \def \showISBNx     #1{\unskip}     \fi
\ifx \showISBNxiii \undefined \def \showISBNxiii  #1{\unskip}     \fi
\ifx \showISSN     \undefined \def \showISSN      #1{\unskip}     \fi
\ifx \showLCCN     \undefined \def \showLCCN      #1{\unskip}     \fi
\ifx \shownote     \undefined \def \shownote      #1{#1}          \fi
\ifx \showarticletitle \undefined \def \showarticletitle #1{#1}   \fi
\ifx \showURL      \undefined \def \showURL       {\relax}        \fi
\providecommand\bibfield[2]{#2}
\providecommand\bibinfo[2]{#2}
\providecommand\natexlab[1]{#1}
\providecommand\showeprint[2][]{arXiv:#2}

\bibitem[Alexander et~al\mbox{.}(2022)]%
        {Alexander2022ORCASIQA}
\bibfield{author}{\bibinfo{person}{Daria Alexander}, \bibinfo{person}{Wojciech Kusa}, {and} \bibinfo{person}{Arjen~P. de Vries}.} \bibinfo{year}{2022}\natexlab{}.
\newblock \showarticletitle{ORCAS-I: Queries Annotated with Intent using Weak Supervision}.
\newblock \bibinfo{journal}{\emph{Proceedings of the 45th International ACM SIGIR Conference on Research and Development in Information Retrieval}} (\bibinfo{year}{2022}).
\newblock
\urldef\tempurl%
\url{https://api.semanticscholar.org/CorpusID:248495926}
\showURL{%
\tempurl}


\bibitem[Ankalkoti(2017)]%
        {Ankalkoti2017SurveyOS}
\bibfield{author}{\bibinfo{person}{Prashant Ankalkoti}.} \bibinfo{year}{2017}\natexlab{}.
\newblock \showarticletitle{Survey on Search Engine Optimization Tools \& Techniques}.
\newblock \bibinfo{journal}{\emph{Imperial journal of interdisciplinary research}}  \bibinfo{volume}{3} (\bibinfo{year}{2017}).
\newblock
\urldef\tempurl%
\url{https://api.semanticscholar.org/CorpusID:116487363}
\showURL{%
\tempurl}


\bibitem[Asai et~al\mbox{.}(2021)]%
        {Asai2021OneQA}
\bibfield{author}{\bibinfo{person}{Akari Asai}, \bibinfo{person}{Xinyan~Velocity Yu}, \bibinfo{person}{Jungo Kasai}, {and} \bibinfo{person}{Hannaneh Hajishirzi}.} \bibinfo{year}{2021}\natexlab{}.
\newblock \showarticletitle{One Question Answering Model for Many Languages with Cross-lingual Dense Passage Retrieval}. In \bibinfo{booktitle}{\emph{Neural Information Processing Systems}}.
\newblock
\urldef\tempurl%
\url{https://api.semanticscholar.org/CorpusID:236428949}
\showURL{%
\tempurl}


\bibitem[Brin and Page(1998)]%
        {Brin1998TheAO}
\bibfield{author}{\bibinfo{person}{Sergey Brin} {and} \bibinfo{person}{Lawrence Page}.} \bibinfo{year}{1998}\natexlab{}.
\newblock \showarticletitle{The Anatomy of a Large-Scale Hypertextual Web Search Engine}.
\newblock \bibinfo{journal}{\emph{Comput. Networks}}  \bibinfo{volume}{30} (\bibinfo{year}{1998}), \bibinfo{pages}{107--117}.
\newblock
\urldef\tempurl%
\url{https://api.semanticscholar.org/CorpusID:7587743}
\showURL{%
\tempurl}


\bibitem[Brown et~al\mbox{.}(2020)]%
        {NEURIPS2020_1457c0d6}
\bibfield{author}{\bibinfo{person}{Tom Brown}, \bibinfo{person}{Benjamin Mann}, \bibinfo{person}{Nick Ryder}, \bibinfo{person}{Melanie Subbiah}, \bibinfo{person}{Jared~D Kaplan}, \bibinfo{person}{Prafulla Dhariwal}, \bibinfo{person}{Arvind Neelakantan}, \bibinfo{person}{Pranav Shyam}, \bibinfo{person}{Girish Sastry}, \bibinfo{person}{Amanda Askell}, \bibinfo{person}{Sandhini Agarwal}, \bibinfo{person}{Ariel Herbert-Voss}, \bibinfo{person}{Gretchen Krueger}, \bibinfo{person}{Tom Henighan}, \bibinfo{person}{Rewon Child}, \bibinfo{person}{Aditya Ramesh}, \bibinfo{person}{Daniel Ziegler}, \bibinfo{person}{Jeffrey Wu}, \bibinfo{person}{Clemens Winter}, \bibinfo{person}{Chris Hesse}, \bibinfo{person}{Mark Chen}, \bibinfo{person}{Eric Sigler}, \bibinfo{person}{Mateusz Litwin}, \bibinfo{person}{Scott Gray}, \bibinfo{person}{Benjamin Chess}, \bibinfo{person}{Jack Clark}, \bibinfo{person}{Christopher Berner}, \bibinfo{person}{Sam McCandlish}, \bibinfo{person}{Alec Radford}, \bibinfo{person}{Ilya Sutskever}, {and}
  \bibinfo{person}{Dario Amodei}.} \bibinfo{year}{2020}\natexlab{}.
\newblock \showarticletitle{Language Models are Few-Shot Learners}. In \bibinfo{booktitle}{\emph{Advances in Neural Information Processing Systems}}, \bibfield{editor}{\bibinfo{person}{H.~Larochelle}, \bibinfo{person}{M.~Ranzato}, \bibinfo{person}{R.~Hadsell}, \bibinfo{person}{M.F. Balcan}, {and} \bibinfo{person}{H.~Lin}} (Eds.), Vol.~\bibinfo{volume}{33}. \bibinfo{publisher}{Curran Associates, Inc.}, \bibinfo{pages}{1877--1901}.
\newblock
\urldef\tempurl%
\url{https://proceedings.neurips.cc/paper_files/paper/2020/file/1457c0d6bfcb4967418bfb8ac142f64a-Paper.pdf}
\showURL{%
\tempurl}


\bibitem[Craswell et~al\mbox{.}(2021)]%
        {Craswell2021MSMB}
\bibfield{author}{\bibinfo{person}{Nick Craswell}, \bibinfo{person}{Bhaskar Mitra}, \bibinfo{person}{Emine Yilmaz}, \bibinfo{person}{Daniel~Fernando Campos}, {and} \bibinfo{person}{Jimmy~J. Lin}.} \bibinfo{year}{2021}\natexlab{}.
\newblock \showarticletitle{MS MARCO: Benchmarking Ranking Models in the Large-Data Regime}.
\newblock \bibinfo{journal}{\emph{Proceedings of the 44th International ACM SIGIR Conference on Research and Development in Information Retrieval}} (\bibinfo{year}{2021}).
\newblock
\urldef\tempurl%
\url{https://api.semanticscholar.org/CorpusID:234336491}
\showURL{%
\tempurl}


\bibitem[Dean(2023)]%
        {dean2023}
\bibfield{author}{\bibinfo{person}{Brian Dean}.} \bibinfo{year}{2023}\natexlab{}.
\newblock \bibinfo{title}{We Analyzed 4 Million Google Search Results. Here's What We Learned About Organic Click Through Rate}.
\newblock
\newblock
\urldef\tempurl%
\url{https://backlinko.com/google-ctr-stats}
\showURL{%
\tempurl}
\newblock
\shownote{Accessed: 2024-06-08}.


\bibitem[Goodwin(2011)]%
        {searchenginewatch2011}
\bibfield{author}{\bibinfo{person}{Danny Goodwin}.} \bibinfo{year}{2011}\natexlab{}.
\newblock \bibinfo{title}{Top Google Result Gets 36.4\% of Clicks [Study]}.
\newblock
\newblock
\urldef\tempurl%
\url{https://www.searchenginewatch.com/2011/04/21/top-google-result-gets-36-4-of-clicks-study/}
\showURL{%
\tempurl}


\bibitem[Guu et~al\mbox{.}(2020)]%
        {Guu2020REALMRL}
\bibfield{author}{\bibinfo{person}{Kelvin Guu}, \bibinfo{person}{Kenton Lee}, \bibinfo{person}{Zora Tung}, \bibinfo{person}{Panupong Pasupat}, {and} \bibinfo{person}{Ming-Wei Chang}.} \bibinfo{year}{2020}\natexlab{}.
\newblock \showarticletitle{REALM: Retrieval-Augmented Language Model Pre-Training}.
\newblock \bibinfo{journal}{\emph{ArXiv}}  \bibinfo{volume}{abs/2002.08909} (\bibinfo{year}{2020}).
\newblock
\urldef\tempurl%
\url{https://api.semanticscholar.org/CorpusID:211204736}
\showURL{%
\tempurl}


\bibitem[Ji et~al\mbox{.}(2023)]%
        {ji2023survey}
\bibfield{author}{\bibinfo{person}{Ziwei Ji}, \bibinfo{person}{Nayeon Lee}, \bibinfo{person}{Rita Frieske}, \bibinfo{person}{Tiezheng Yu}, \bibinfo{person}{Dan Su}, \bibinfo{person}{Yan Xu}, \bibinfo{person}{Etsuko Ishii}, \bibinfo{person}{Ye~Jin Bang}, \bibinfo{person}{Andrea Madotto}, {and} \bibinfo{person}{Pascale Fung}.} \bibinfo{year}{2023}\natexlab{}.
\newblock \showarticletitle{Survey of hallucination in natural language generation}.
\newblock \bibinfo{journal}{\emph{Comput. Surveys}} \bibinfo{volume}{55}, \bibinfo{number}{12} (\bibinfo{year}{2023}), \bibinfo{pages}{1--38}.
\newblock


\bibitem[Kumar and Lakkaraju(2024)]%
        {kumar2024manipulating}
\bibfield{author}{\bibinfo{person}{Aounon Kumar} {and} \bibinfo{person}{Himabindu Lakkaraju}.} \bibinfo{year}{2024}\natexlab{}.
\newblock \bibinfo{title}{Manipulating Large Language Models to Increase Product Visibility}.
\newblock
\newblock
\showeprint[arxiv]{2404.07981}~[cs.IR]


\bibitem[Kumar et~al\mbox{.}(2019)]%
        {Kumar2019ASO}
\bibfield{author}{\bibinfo{person}{R.Anil Kumar}, \bibinfo{person}{Zaiduddin Shaik}, {and} \bibinfo{person}{Mohammed Furqan}.} \bibinfo{year}{2019}\natexlab{}.
\newblock \showarticletitle{A Survey on Search Engine Optimization Techniques}.
\newblock \bibinfo{journal}{\emph{International Journal of P2P Network Trends and Technology}} (\bibinfo{year}{2019}).
\newblock
\urldef\tempurl%
\url{https://doi.org/10.14445/22492615/IJPTT-V9I1P402}
\showDOI{\tempurl}


\bibitem[Kwiatkowski et~al\mbox{.}(2019)]%
        {Kwiatkowski2019NaturalQA}
\bibfield{author}{\bibinfo{person}{Tom Kwiatkowski}, \bibinfo{person}{Jennimaria Palomaki}, \bibinfo{person}{Olivia Redfield}, \bibinfo{person}{Michael Collins}, \bibinfo{person}{Ankur~P. Parikh}, \bibinfo{person}{Chris Alberti}, \bibinfo{person}{Danielle Epstein}, \bibinfo{person}{Illia Polosukhin}, \bibinfo{person}{Jacob Devlin}, \bibinfo{person}{Kenton Lee}, \bibinfo{person}{Kristina Toutanova}, \bibinfo{person}{Llion Jones}, \bibinfo{person}{Matthew Kelcey}, \bibinfo{person}{Ming-Wei Chang}, \bibinfo{person}{Andrew~M. Dai}, \bibinfo{person}{Jakob Uszkoreit}, \bibinfo{person}{Quoc~V. Le}, {and} \bibinfo{person}{Slav Petrov}.} \bibinfo{year}{2019}\natexlab{}.
\newblock \showarticletitle{Natural Questions: A Benchmark for Question Answering Research}.
\newblock \bibinfo{journal}{\emph{Transactions of the Association for Computational Linguistics}}  \bibinfo{volume}{7} (\bibinfo{year}{2019}), \bibinfo{pages}{453--466}.
\newblock
\urldef\tempurl%
\url{https://api.semanticscholar.org/CorpusID:86611921}
\showURL{%
\tempurl}


\bibitem[Liu et~al\mbox{.}(2023b)]%
        {Liu2023EvaluatingVI}
\bibfield{author}{\bibinfo{person}{Nelson~F. Liu}, \bibinfo{person}{Tianyi Zhang}, {and} \bibinfo{person}{Percy Liang}.} \bibinfo{year}{2023}\natexlab{b}.
\newblock \showarticletitle{Evaluating Verifiability in Generative Search Engines}.
\newblock \bibinfo{journal}{\emph{ArXiv}}  \bibinfo{volume}{abs/2304.09848} (\bibinfo{year}{2023}).
\newblock
\urldef\tempurl%
\url{https://api.semanticscholar.org/CorpusID:258212854}
\showURL{%
\tempurl}


\bibitem[Liu et~al\mbox{.}(2023a)]%
        {Liu2023GEvalNE}
\bibfield{author}{\bibinfo{person}{Yang Liu}, \bibinfo{person}{Dan Iter}, \bibinfo{person}{Yichong Xu}, \bibinfo{person}{Shuo Wang}, \bibinfo{person}{Ruochen Xu}, {and} \bibinfo{person}{Chenguang Zhu}.} \bibinfo{year}{2023}\natexlab{a}.
\newblock \showarticletitle{G-Eval: NLG Evaluation using GPT-4 with Better Human Alignment}.
\newblock \bibinfo{journal}{\emph{ArXiv}}  \bibinfo{volume}{abs/2303.16634} (\bibinfo{year}{2023}).
\newblock
\urldef\tempurl%
\url{https://api.semanticscholar.org/CorpusID:257804696}
\showURL{%
\tempurl}


\bibitem[Maayan(2023)]%
        {Maayan2023}
\bibfield{author}{\bibinfo{person}{G.~D. Maayan}.} \bibinfo{year}{2023}\natexlab{}.
\newblock \showarticletitle{How Google SGE will impact your traffic – and 3 SGE recovery case studies}.
\newblock \bibinfo{journal}{\emph{Search Engine Land}} (\bibinfo{date}{5 Sep} \bibinfo{year}{2023}).
\newblock
\urldef\tempurl%
\url{https://searchengineland.com/how-google-sge-will-impact-your-traffic-and-3-sge-recovery-case-studies-431430}
\showURL{%
\tempurl}


\bibitem[Menick et~al\mbox{.}(2022)]%
        {Menick2022TeachingLM}
\bibfield{author}{\bibinfo{person}{Jacob Menick}, \bibinfo{person}{Maja Trebacz}, \bibinfo{person}{Vladimir Mikulik}, \bibinfo{person}{John Aslanides}, \bibinfo{person}{Francis Song}, \bibinfo{person}{Martin Chadwick}, \bibinfo{person}{Mia Glaese}, \bibinfo{person}{Susannah Young}, \bibinfo{person}{Lucy Campbell-Gillingham}, \bibinfo{person}{Geoffrey Irving}, {and} \bibinfo{person}{Nathan McAleese}.} \bibinfo{year}{2022}\natexlab{}.
\newblock \showarticletitle{Teaching language models to support answers with verified quotes}.
\newblock \bibinfo{journal}{\emph{ArXiv}}  \bibinfo{volume}{abs/2203.11147} (\bibinfo{year}{2022}).
\newblock
\urldef\tempurl%
\url{https://api.semanticscholar.org/CorpusID:247594830}
\showURL{%
\tempurl}


\bibitem[Mialon et~al\mbox{.}(2023)]%
        {Mialon2023AugmentedLM}
\bibfield{author}{\bibinfo{person}{Gr{\'e}goire Mialon}, \bibinfo{person}{Roberto Dess{\`i}}, \bibinfo{person}{Maria Lomeli}, \bibinfo{person}{Christoforos Nalmpantis}, \bibinfo{person}{Ramakanth Pasunuru}, \bibinfo{person}{Roberta Raileanu}, \bibinfo{person}{Baptiste Rozi{\`e}re}, \bibinfo{person}{Timo Schick}, \bibinfo{person}{Jane Dwivedi-Yu}, \bibinfo{person}{Asli Celikyilmaz}, \bibinfo{person}{Edouard Grave}, \bibinfo{person}{Yann LeCun}, {and} \bibinfo{person}{Thomas Scialom}.} \bibinfo{year}{2023}\natexlab{}.
\newblock \showarticletitle{Augmented Language Models: a Survey}.
\newblock \bibinfo{journal}{\emph{ArXiv}}  \bibinfo{volume}{abs/2302.07842} (\bibinfo{year}{2023}).
\newblock
\urldef\tempurl%
\url{https://api.semanticscholar.org/CorpusID:256868474}
\showURL{%
\tempurl}


\bibitem[Nakano et~al\mbox{.}(2021)]%
        {Nakano2021WebGPTBQ}
\bibfield{author}{\bibinfo{person}{Reiichiro Nakano}, \bibinfo{person}{Jacob Hilton}, \bibinfo{person}{S.~Arun Balaji}, \bibinfo{person}{Jeff Wu}, \bibinfo{person}{Ouyang Long}, \bibinfo{person}{Christina Kim}, \bibinfo{person}{Christopher Hesse}, \bibinfo{person}{Shantanu Jain}, \bibinfo{person}{Vineet Kosaraju}, \bibinfo{person}{William Saunders}, \bibinfo{person}{Xu Jiang}, \bibinfo{person}{Karl Cobbe}, \bibinfo{person}{Tyna Eloundou}, \bibinfo{person}{Gretchen Krueger}, \bibinfo{person}{Kevin Button}, \bibinfo{person}{Matthew Knight}, \bibinfo{person}{Benjamin Chess}, {and} \bibinfo{person}{John Schulman}.} \bibinfo{year}{2021}\natexlab{}.
\newblock \showarticletitle{WebGPT: Browser-assisted question-answering with human feedback}.
\newblock \bibinfo{journal}{\emph{ArXiv}}  \bibinfo{volume}{abs/2112.09332} (\bibinfo{year}{2021}).
\newblock
\urldef\tempurl%
\url{https://api.semanticscholar.org/CorpusID:245329531}
\showURL{%
\tempurl}


\bibitem[OpenAI(2022)]%
        {gpt35turbo}
\bibfield{author}{\bibinfo{person}{OpenAI}.} \bibinfo{year}{2022}\natexlab{}.
\newblock \bibinfo{title}{Introducing ChatGPT}.
\newblock
\newblock
\urldef\tempurl%
\url{https://openai.com/index/chatgpt/}
\showURL{%
\tempurl}


\bibitem[OpenAI et~al\mbox{.}(2024)]%
        {openai2024gpt4}
\bibfield{author}{\bibinfo{person}{OpenAI}, \bibinfo{person}{Josh Achiam}, \bibinfo{person}{Steven Adler}, \bibinfo{person}{Sandhini Agarwal}, \bibinfo{person}{Lama Ahmad}, \bibinfo{person}{Ilge Akkaya}, \bibinfo{person}{Florencia~Leoni Aleman}, \bibinfo{person}{Diogo Almeida}, \bibinfo{person}{Janko Altenschmidt}, \bibinfo{person}{Sam Altman}, \bibinfo{person}{Shyamal Anadkat}, \bibinfo{person}{Red Avila}, \bibinfo{person}{Igor Babuschkin}, \bibinfo{person}{Suchir Balaji}, \bibinfo{person}{Valerie Balcom}, \bibinfo{person}{Paul Baltescu}, \bibinfo{person}{Haiming Bao}, \bibinfo{person}{Mohammad Bavarian}, \bibinfo{person}{Jeff Belgum}, \bibinfo{person}{Irwan Bello}, \bibinfo{person}{Jake Berdine}, \bibinfo{person}{Gabriel Bernadett-Shapiro}, \bibinfo{person}{Christopher Berner}, \bibinfo{person}{Lenny Bogdonoff}, \bibinfo{person}{Oleg Boiko}, \bibinfo{person}{Madelaine Boyd}, \bibinfo{person}{Anna-Luisa Brakman}, \bibinfo{person}{Greg Brockman}, \bibinfo{person}{Tim Brooks}, \bibinfo{person}{Miles Brundage},
  \bibinfo{person}{Kevin Button}, \bibinfo{person}{Trevor Cai}, \bibinfo{person}{Rosie Campbell}, \bibinfo{person}{Andrew Cann}, \bibinfo{person}{Brittany Carey}, \bibinfo{person}{Chelsea Carlson}, \bibinfo{person}{Rory Carmichael}, \bibinfo{person}{Brooke Chan}, \bibinfo{person}{Che Chang}, \bibinfo{person}{Fotis Chantzis}, \bibinfo{person}{Derek Chen}, \bibinfo{person}{Sully Chen}, \bibinfo{person}{Ruby Chen}, \bibinfo{person}{Jason Chen}, \bibinfo{person}{Mark Chen}, \bibinfo{person}{Ben Chess}, \bibinfo{person}{Chester Cho}, \bibinfo{person}{Casey Chu}, \bibinfo{person}{Hyung~Won Chung}, \bibinfo{person}{Dave Cummings}, \bibinfo{person}{Jeremiah Currier}, \bibinfo{person}{Yunxing Dai}, \bibinfo{person}{Cory Decareaux}, \bibinfo{person}{Thomas Degry}, \bibinfo{person}{Noah Deutsch}, \bibinfo{person}{Damien Deville}, \bibinfo{person}{Arka Dhar}, \bibinfo{person}{David Dohan}, \bibinfo{person}{Steve Dowling}, \bibinfo{person}{Sheila Dunning}, \bibinfo{person}{Adrien Ecoffet}, \bibinfo{person}{Atty Eleti},
  \bibinfo{person}{Tyna Eloundou}, \bibinfo{person}{David Farhi}, \bibinfo{person}{Liam Fedus}, \bibinfo{person}{Niko Felix}, \bibinfo{person}{Simón~Posada Fishman}, \bibinfo{person}{Juston Forte}, \bibinfo{person}{Isabella Fulford}, \bibinfo{person}{Leo Gao}, \bibinfo{person}{Elie Georges}, \bibinfo{person}{Christian Gibson}, \bibinfo{person}{Vik Goel}, \bibinfo{person}{Tarun Gogineni}, \bibinfo{person}{Gabriel Goh}, \bibinfo{person}{Rapha Gontijo-Lopes}, \bibinfo{person}{Jonathan Gordon}, \bibinfo{person}{Morgan Grafstein}, \bibinfo{person}{Scott Gray}, \bibinfo{person}{Ryan Greene}, \bibinfo{person}{Joshua Gross}, \bibinfo{person}{Shixiang~Shane Gu}, \bibinfo{person}{Yufei Guo}, \bibinfo{person}{Chris Hallacy}, \bibinfo{person}{Jesse Han}, \bibinfo{person}{Jeff Harris}, \bibinfo{person}{Yuchen He}, \bibinfo{person}{Mike Heaton}, \bibinfo{person}{Johannes Heidecke}, \bibinfo{person}{Chris Hesse}, \bibinfo{person}{Alan Hickey}, \bibinfo{person}{Wade Hickey}, \bibinfo{person}{Peter Hoeschele},
  \bibinfo{person}{Brandon Houghton}, \bibinfo{person}{Kenny Hsu}, \bibinfo{person}{Shengli Hu}, \bibinfo{person}{Xin Hu}, \bibinfo{person}{Joost Huizinga}, \bibinfo{person}{Shantanu Jain}, \bibinfo{person}{Shawn Jain}, \bibinfo{person}{Joanne Jang}, \bibinfo{person}{Angela Jiang}, \bibinfo{person}{Roger Jiang}, \bibinfo{person}{Haozhun Jin}, \bibinfo{person}{Denny Jin}, \bibinfo{person}{Shino Jomoto}, \bibinfo{person}{Billie Jonn}, \bibinfo{person}{Heewoo Jun}, \bibinfo{person}{Tomer Kaftan}, \bibinfo{person}{Łukasz Kaiser}, \bibinfo{person}{Ali Kamali}, \bibinfo{person}{Ingmar Kanitscheider}, \bibinfo{person}{Nitish~Shirish Keskar}, \bibinfo{person}{Tabarak Khan}, \bibinfo{person}{Logan Kilpatrick}, \bibinfo{person}{Jong~Wook Kim}, \bibinfo{person}{Christina Kim}, \bibinfo{person}{Yongjik Kim}, \bibinfo{person}{Jan~Hendrik Kirchner}, \bibinfo{person}{Jamie Kiros}, \bibinfo{person}{Matt Knight}, \bibinfo{person}{Daniel Kokotajlo}, \bibinfo{person}{Łukasz Kondraciuk}, \bibinfo{person}{Andrew Kondrich},
  \bibinfo{person}{Aris Konstantinidis}, \bibinfo{person}{Kyle Kosic}, \bibinfo{person}{Gretchen Krueger}, \bibinfo{person}{Vishal Kuo}, \bibinfo{person}{Michael Lampe}, \bibinfo{person}{Ikai Lan}, \bibinfo{person}{Teddy Lee}, \bibinfo{person}{Jan Leike}, \bibinfo{person}{Jade Leung}, \bibinfo{person}{Daniel Levy}, \bibinfo{person}{Chak~Ming Li}, \bibinfo{person}{Rachel Lim}, \bibinfo{person}{Molly Lin}, \bibinfo{person}{Stephanie Lin}, \bibinfo{person}{Mateusz Litwin}, \bibinfo{person}{Theresa Lopez}, \bibinfo{person}{Ryan Lowe}, \bibinfo{person}{Patricia Lue}, \bibinfo{person}{Anna Makanju}, \bibinfo{person}{Kim Malfacini}, \bibinfo{person}{Sam Manning}, \bibinfo{person}{Todor Markov}, \bibinfo{person}{Yaniv Markovski}, \bibinfo{person}{Bianca Martin}, \bibinfo{person}{Katie Mayer}, \bibinfo{person}{Andrew Mayne}, \bibinfo{person}{Bob McGrew}, \bibinfo{person}{Scott~Mayer McKinney}, \bibinfo{person}{Christine McLeavey}, \bibinfo{person}{Paul McMillan}, \bibinfo{person}{Jake McNeil}, \bibinfo{person}{David
  Medina}, \bibinfo{person}{Aalok Mehta}, \bibinfo{person}{Jacob Menick}, \bibinfo{person}{Luke Metz}, \bibinfo{person}{Andrey Mishchenko}, \bibinfo{person}{Pamela Mishkin}, \bibinfo{person}{Vinnie Monaco}, \bibinfo{person}{Evan Morikawa}, \bibinfo{person}{Daniel Mossing}, \bibinfo{person}{Tong Mu}, \bibinfo{person}{Mira Murati}, \bibinfo{person}{Oleg Murk}, \bibinfo{person}{David Mély}, \bibinfo{person}{Ashvin Nair}, \bibinfo{person}{Reiichiro Nakano}, \bibinfo{person}{Rajeev Nayak}, \bibinfo{person}{Arvind Neelakantan}, \bibinfo{person}{Richard Ngo}, \bibinfo{person}{Hyeonwoo Noh}, \bibinfo{person}{Long Ouyang}, \bibinfo{person}{Cullen O'Keefe}, \bibinfo{person}{Jakub Pachocki}, \bibinfo{person}{Alex Paino}, \bibinfo{person}{Joe Palermo}, \bibinfo{person}{Ashley Pantuliano}, \bibinfo{person}{Giambattista Parascandolo}, \bibinfo{person}{Joel Parish}, \bibinfo{person}{Emy Parparita}, \bibinfo{person}{Alex Passos}, \bibinfo{person}{Mikhail Pavlov}, \bibinfo{person}{Andrew Peng}, \bibinfo{person}{Adam
  Perelman}, \bibinfo{person}{Filipe de Avila Belbute~Peres}, \bibinfo{person}{Michael Petrov}, \bibinfo{person}{Henrique~Ponde de Oliveira~Pinto}, \bibinfo{person}{Michael}, \bibinfo{person}{Pokorny}, \bibinfo{person}{Michelle Pokrass}, \bibinfo{person}{Vitchyr~H. Pong}, \bibinfo{person}{Tolly Powell}, \bibinfo{person}{Alethea Power}, \bibinfo{person}{Boris Power}, \bibinfo{person}{Elizabeth Proehl}, \bibinfo{person}{Raul Puri}, \bibinfo{person}{Alec Radford}, \bibinfo{person}{Jack Rae}, \bibinfo{person}{Aditya Ramesh}, \bibinfo{person}{Cameron Raymond}, \bibinfo{person}{Francis Real}, \bibinfo{person}{Kendra Rimbach}, \bibinfo{person}{Carl Ross}, \bibinfo{person}{Bob Rotsted}, \bibinfo{person}{Henri Roussez}, \bibinfo{person}{Nick Ryder}, \bibinfo{person}{Mario Saltarelli}, \bibinfo{person}{Ted Sanders}, \bibinfo{person}{Shibani Santurkar}, \bibinfo{person}{Girish Sastry}, \bibinfo{person}{Heather Schmidt}, \bibinfo{person}{David Schnurr}, \bibinfo{person}{John Schulman}, \bibinfo{person}{Daniel Selsam},
  \bibinfo{person}{Kyla Sheppard}, \bibinfo{person}{Toki Sherbakov}, \bibinfo{person}{Jessica Shieh}, \bibinfo{person}{Sarah Shoker}, \bibinfo{person}{Pranav Shyam}, \bibinfo{person}{Szymon Sidor}, \bibinfo{person}{Eric Sigler}, \bibinfo{person}{Maddie Simens}, \bibinfo{person}{Jordan Sitkin}, \bibinfo{person}{Katarina Slama}, \bibinfo{person}{Ian Sohl}, \bibinfo{person}{Benjamin Sokolowsky}, \bibinfo{person}{Yang Song}, \bibinfo{person}{Natalie Staudacher}, \bibinfo{person}{Felipe~Petroski Such}, \bibinfo{person}{Natalie Summers}, \bibinfo{person}{Ilya Sutskever}, \bibinfo{person}{Jie Tang}, \bibinfo{person}{Nikolas Tezak}, \bibinfo{person}{Madeleine~B. Thompson}, \bibinfo{person}{Phil Tillet}, \bibinfo{person}{Amin Tootoonchian}, \bibinfo{person}{Elizabeth Tseng}, \bibinfo{person}{Preston Tuggle}, \bibinfo{person}{Nick Turley}, \bibinfo{person}{Jerry Tworek}, \bibinfo{person}{Juan Felipe~Cerón Uribe}, \bibinfo{person}{Andrea Vallone}, \bibinfo{person}{Arun Vijayvergiya}, \bibinfo{person}{Chelsea Voss},
  \bibinfo{person}{Carroll Wainwright}, \bibinfo{person}{Justin~Jay Wang}, \bibinfo{person}{Alvin Wang}, \bibinfo{person}{Ben Wang}, \bibinfo{person}{Jonathan Ward}, \bibinfo{person}{Jason Wei}, \bibinfo{person}{CJ Weinmann}, \bibinfo{person}{Akila Welihinda}, \bibinfo{person}{Peter Welinder}, \bibinfo{person}{Jiayi Weng}, \bibinfo{person}{Lilian Weng}, \bibinfo{person}{Matt Wiethoff}, \bibinfo{person}{Dave Willner}, \bibinfo{person}{Clemens Winter}, \bibinfo{person}{Samuel Wolrich}, \bibinfo{person}{Hannah Wong}, \bibinfo{person}{Lauren Workman}, \bibinfo{person}{Sherwin Wu}, \bibinfo{person}{Jeff Wu}, \bibinfo{person}{Michael Wu}, \bibinfo{person}{Kai Xiao}, \bibinfo{person}{Tao Xu}, \bibinfo{person}{Sarah Yoo}, \bibinfo{person}{Kevin Yu}, \bibinfo{person}{Qiming Yuan}, \bibinfo{person}{Wojciech Zaremba}, \bibinfo{person}{Rowan Zellers}, \bibinfo{person}{Chong Zhang}, \bibinfo{person}{Marvin Zhang}, \bibinfo{person}{Shengjia Zhao}, \bibinfo{person}{Tianhao Zheng}, \bibinfo{person}{Juntang Zhuang},
  \bibinfo{person}{William Zhuk}, {and} \bibinfo{person}{Barret Zoph}.} \bibinfo{year}{2024}\natexlab{}.
\newblock \bibinfo{title}{GPT-4 Technical Report}.
\newblock
\newblock
\showeprint[arxiv]{2303.08774}~[cs.CL]


\bibitem[Shahzad et~al\mbox{.}(2020)]%
        {Shahzad2020TheNT}
\bibfield{author}{\bibinfo{person}{A. Shahzad}, \bibinfo{person}{Deden~Witarsyah Jacob}, \bibinfo{person}{Nazri~M. Nawi}, \bibinfo{person}{Hairulnizam~Bin Mahdin}, {and} \bibinfo{person}{Marheni~Eka Saputri}.} \bibinfo{year}{2020}\natexlab{}.
\newblock \showarticletitle{The new trend for search engine optimization, tools and techniques}.
\newblock \bibinfo{journal}{\emph{Indonesian Journal of Electrical Engineering and Computer Science}}  \bibinfo{volume}{18} (\bibinfo{year}{2020}), \bibinfo{pages}{1568}.
\newblock
\urldef\tempurl%
\url{https://api.semanticscholar.org/CorpusID:213123106}
\showURL{%
\tempurl}


\bibitem[Shuster et~al\mbox{.}(2022)]%
        {Shuster2022BlenderBot3A}
\bibfield{author}{\bibinfo{person}{Kurt Shuster}, \bibinfo{person}{Jing Xu}, \bibinfo{person}{Mojtaba Komeili}, \bibinfo{person}{Da Ju}, \bibinfo{person}{Eric~Michael Smith}, \bibinfo{person}{Stephen Roller}, \bibinfo{person}{Megan Ung}, \bibinfo{person}{Moya Chen}, \bibinfo{person}{Kushal Arora}, \bibinfo{person}{Joshua Lane}, \bibinfo{person}{Morteza Behrooz}, \bibinfo{person}{W.K.F. Ngan}, \bibinfo{person}{Spencer Poff}, \bibinfo{person}{Naman Goyal}, \bibinfo{person}{Arthur Szlam}, \bibinfo{person}{Y-Lan Boureau}, \bibinfo{person}{Melanie Kambadur}, {and} \bibinfo{person}{Jason Weston}.} \bibinfo{year}{2022}\natexlab{}.
\newblock \showarticletitle{BlenderBot 3: a deployed conversational agent that continually learns to responsibly engage}.
\newblock \bibinfo{journal}{\emph{ArXiv}}  \bibinfo{volume}{abs/2208.03188} (\bibinfo{year}{2022}).
\newblock
\urldef\tempurl%
\url{https://api.semanticscholar.org/CorpusID:251371589}
\showURL{%
\tempurl}


\bibitem[Thoppilan et~al\mbox{.}(2022)]%
        {Thoppilan2022LaMDALM}
\bibfield{author}{\bibinfo{person}{Romal Thoppilan}, \bibinfo{person}{Daniel~De Freitas}, \bibinfo{person}{Jamie Hall}, \bibinfo{person}{Noam Shazeer}, \bibinfo{person}{Apoorv Kulshreshtha}, \bibinfo{person}{Heng-Tze Cheng}, \bibinfo{person}{Alicia Jin}, \bibinfo{person}{Taylor Bos}, \bibinfo{person}{Leslie Baker}, \bibinfo{person}{Yu Du}, \bibinfo{person}{YaGuang Li}, \bibinfo{person}{Hongrae Lee}, \bibinfo{person}{Huaixiu~Steven Zheng}, \bibinfo{person}{Amin Ghafouri}, \bibinfo{person}{Marcelo Menegali}, \bibinfo{person}{Yanping Huang}, \bibinfo{person}{Maxim Krikun}, \bibinfo{person}{Dmitry Lepikhin}, \bibinfo{person}{James Qin}, \bibinfo{person}{Dehao Chen}, \bibinfo{person}{Yuanzhong Xu}, \bibinfo{person}{Zhifeng Chen}, \bibinfo{person}{Adam Roberts}, \bibinfo{person}{Maarten Bosma}, \bibinfo{person}{Vincent Zhao}, \bibinfo{person}{Yanqi Zhou}, \bibinfo{person}{Chung-Ching Chang}, \bibinfo{person}{Igor Krivokon}, \bibinfo{person}{Will Rusch}, \bibinfo{person}{Marc Pickett}, \bibinfo{person}{Pranesh
  Srinivasan}, \bibinfo{person}{Laichee Man}, \bibinfo{person}{Kathleen Meier-Hellstern}, \bibinfo{person}{Meredith~Ringel Morris}, \bibinfo{person}{Tulsee Doshi}, \bibinfo{person}{Renelito~Delos Santos}, \bibinfo{person}{Toju Duke}, \bibinfo{person}{Johnny Soraker}, \bibinfo{person}{Ben Zevenbergen}, \bibinfo{person}{Vinodkumar Prabhakaran}, \bibinfo{person}{Mark Diaz}, \bibinfo{person}{Ben Hutchinson}, \bibinfo{person}{Kristen Olson}, \bibinfo{person}{Alejandra Molina}, \bibinfo{person}{Erin Hoffman-John}, \bibinfo{person}{Josh Lee}, \bibinfo{person}{Lora Aroyo}, \bibinfo{person}{Ravi Rajakumar}, \bibinfo{person}{Alena Butryna}, \bibinfo{person}{Matthew Lamm}, \bibinfo{person}{Viktoriya Kuzmina}, \bibinfo{person}{Joe Fenton}, \bibinfo{person}{Aaron Cohen}, \bibinfo{person}{Rachel Bernstein}, \bibinfo{person}{Ray Kurzweil}, \bibinfo{person}{Blaise Aguera-Arcas}, \bibinfo{person}{Claire Cui}, \bibinfo{person}{Marian Croak}, \bibinfo{person}{Ed Chi}, {and} \bibinfo{person}{Quoc Le}.}
  \bibinfo{year}{2022}\natexlab{}.
\newblock \bibinfo{title}{LaMDA: Language Models for Dialog Applications}.
\newblock
\newblock
\showeprint[arxiv]{2201.08239}~[cs.CL]


\bibitem[Zhou et~al\mbox{.}(2023)]%
        {Zhou2023LIMALI}
\bibfield{author}{\bibinfo{person}{Chunting Zhou}, \bibinfo{person}{Pengfei Liu}, \bibinfo{person}{Puxin Xu}, \bibinfo{person}{Srini Iyer}, \bibinfo{person}{Jiao Sun}, \bibinfo{person}{Yuning Mao}, \bibinfo{person}{Xuezhe Ma}, \bibinfo{person}{Avia Efrat}, \bibinfo{person}{Ping Yu}, \bibinfo{person}{L. Yu}, \bibinfo{person}{Susan Zhang}, \bibinfo{person}{Gargi Ghosh}, \bibinfo{person}{Mike Lewis}, \bibinfo{person}{Luke Zettlemoyer}, {and} \bibinfo{person}{Omer Levy}.} \bibinfo{year}{2023}\natexlab{}.
\newblock \showarticletitle{LIMA: Less Is More for Alignment}.
\newblock \bibinfo{journal}{\emph{ArXiv}}  \bibinfo{volume}{abs/2305.11206} (\bibinfo{year}{2023}).
\newblock
\urldef\tempurl%
\url{https://api.semanticscholar.org/CorpusID:258822910}
\showURL{%
\tempurl}


\end{thebibliography}

\clearpage
\appendix

\section{Conversational \Le{}}
\label{app:conversationaL_le}

In Section~\ref{sec:formulation_le}, we discussed a single-turn \Le that outputs a single response given the user query. 
However, one of the strengths of upcoming \Le{}s will be their ability to engage in an active back-and-forth conversation with the user.
The conversation allows users to provide clarifications to their queries or \Le{} response and ask follow-ups. Specifically, in equation~\ref{eq:le_single_turn}, instead of the input being a single query $q_u$, it is modeled as a conversation history $H = (q_{u}^{t}, r^{t})$ pairs. 
The response $r^{t+1}$ is then defined as:
\begin{equation}
\label{eq:app_le_conversation}
 GE := f_{LE}(H, P_{U}) \rightarrow r^{t+1}
\end{equation}
where $t$ is the turn number.

Further, to engage the user in a conversation, a separate LLM, $L_{follow}$ or $L_{resp}$, may generate suggested follow-up queries based on $H$, $P_U$, and $r^{t+1}$. 
The suggested follow-up queries are typically designed to maximize the likelihood of user engagement. This not only benefits \Le{} providers by increasing user interaction but also benefits website owners by enhancing their visibility. Furthermore, these follow-up queries can help users by getting more detailed information.
\section{Experimental Setup}
\label{app:exp_setup_le}

\subsection{Evaluated \Le{}}

The exact prompt used is shown in Listing~\ref{fig:ge_prompt}.

\begin{figure}
\begin{lstlisting}[language=markdown, caption={Prompt used for \Le{}. The \LE{} takes the query and 5 sources as input and outputs the response to query with response grounded in the sources.}, label=fig:ge_prompt]
Write an accurate and concise answer for the given user question, using _only_ the provided summarized web search results. The answer should be correct, high-quality, and written by an expert using an unbiased and journalistic tone. The user's language of choice such as English, Francais, Espamol, Deutsch, or  should be used. The answer should be informative, interesting, and engaging. The answer's logic and reasoning should be rigorous and defensible. Every sentence in the answer should be _immediately followed_ by an in-line citation to the search result(s). The cited search result(s) should fully support _all_ the information in the sentence. Search results need to be cited using [index]. When citing several search results, use [1][2][3] format rather than [1, 2, 3]. You can use multiple search results to respond comprehensively while avoiding irrelevant search results.

Question: {query}

Search Results:
{source_text}
\end{lstlisting}
\end{figure}
\begin{figure}
\begin{lstlisting}[language=markdown, caption={Representative Queries from each of the 9 datasets in \bench{}}, label=fig:queries]
### ORCAS
- what does globalization mean
- wine pairing list

### AllSouls
- Are open-access journals the future of academic publishing?
- Should the study of non-Western philosophy be a requirement for a philosophy degree in the UK?

### Davinci-Debate
- Should all citizens receive a basic income?
- Should governments promote atheism?

### ELI5
- Why does my cat kick its toys when playing with them?
- what does caffeine actually do your muscles, especially regarding exercising?

### GPT-4
- What are the benefits of a keto diet?
- What are the most profound impacts of the Renaissance period on modern society?

### LIMA
- What are the primary factors that influence consumer behavior?
- What would be a great twist for a murder mystery? I'm looking for something creative, not to rehash old tropes.

### MS-Macro
- what does monogamous
- what is the normal fbs range for children

### Natural Questions
- where does the phrase bee line come from
- what is the prince of persia in the bible

### Perplexity.ai
- how to gain more followers on LinkedIn
- why is blood sugar higher after a meal
\end{lstlisting}
\end{figure}
\subsection{Benchmark}
\label{app:exp_setup_bench}

\textbf{\bench{}} contains queries from nine datasets. Representative queries from each of the datasets are shown in Figure~\ref{fig:queries}. Further, we tag each of the queries based on a pool of 7 different categories. For tagging, we use the GPT-4 model and manually confirm high recall and precision in tagging. However, owing to such an automated system, the tags can be noisy and should not be considered carefully. Details about each of these queries are presented here:

\begin{itemize}[leftmargin=*]
\item \textbf{Difficulty Level:} The complexity of the query, ranging from simple to complex.

\item \textbf{Nature of Query:} The type of information sought by the query, such as factual, opinion, or comparison.

\item \textbf{Genre:} The category or domain of the query, such as arts and entertainment, finance, or science.

\item \textbf{Specific Topics:} The specific subject matter of the query, such as physics, economics, or computer science.

\item \textbf{Sensitivity:} Whether the query involves sensitive topics or not.

\item \textbf{User Intent:} The purpose behind the user's query, such as research, purchase, or entertainment.

\item \textbf{Answer Type:} The format of the answer that the query is seeking, such as fact, opinion, or list.
\end{itemize}

\subsection{Evaluation Metrics}
\label{app:impression}

We use 7 different subjective impression metrics, whose prompts are presented in our our public repository: \url{https://github.com/GEO-optim/GEO}.

\begin{table*}[!h]
\centering
\resizebox{\linewidth}{!}{%
\begin{tabular}{lccccccccccc}
\toprule
\multirow{2}{*}[-6pt]{\large{\bf{Method}}} & \multicolumn{3}{c}{\bf{\wordposmetric}}  & \multicolumn{8}{c}{\bf{\subjectiveimpression}} \\
\cmidrule(lr){2-4} \cmidrule(lr){5-12}
& Word & Position & Overall & Rel. & Infl. & Unique  & Div. & FollowUp & Pos. & Count & Average
\\
\midrule
\rowcolor{gray!20} \multicolumn{12}{c}{Performance without \leo{}} \\ \midrule
\bf{\leobaseline{}} & $19.7$\tablestd{0.7} & $19.6$\tablestd{0.5} & $19.8$\tablestd{0.6} & $19.8$\tablestd{0.9} & $19.8$\tablestd{1.6} & $19.8$\tablestd{0.6} & $19.8$\tablestd{1.1} & $19.8$\tablestd{1.0} & $19.8$\tablestd{1.0} & $19.8$\tablestd{0.9} & $19.8$\tablestd{0.9} \\

\midrule
\rowcolor{gray!20} \multicolumn{12}{c}{Non-Performing \leo{} methods} \\ \midrule
\bf{\leoseo{}} & $19.6$\tablestd{0.5} & $19.5$\tablestd{0.6} & $19.8$\tablestd{0.5} & $20.8$\tablestd{0.8} & $19.8$\tablestd{1.0} & $20.4$\tablestd{0.5} & $20.6$\tablestd{0.9} & $19.9$\tablestd{0.9} & $21.1$\tablestd{1.0} & $21.0$\tablestd{0.9} & $20.6$\tablestd{0.7} \\
\bf{\leounique{}} & $20.6$\tablestd{0.6} & $20.5$\tablestd{0.7} & $20.7$\tablestd{0.5} & $20.8$\tablestd{0.7} & $20.3$\tablestd{1.3} & $20.5$\tablestd{0.3} & $20.9$\tablestd{0.3} & $20.4$\tablestd{0.7} & $21.5$\tablestd{0.6} & $21.2$\tablestd{0.4} & $20.9$\tablestd{0.4} \\

\midrule
\rowcolor{gray!20} \multicolumn{12}{c}{High-Performing \leo{} methods} \\ \midrule
\bf{\leosimple{}} & $21.5$\tablestd{0.7} & $22.0$\tablestd{0.8} & $21.5$\tablestd{0.6} & $21.0$\tablestd{1.1} & $21.1$\tablestd{1.8} & $21.2$\tablestd{0.9} & $20.9$\tablestd{1.1} & $20.6$\tablestd{1.0} & $21.9$\tablestd{1.1} & $21.4$\tablestd{0.9} & $21.3$\tablestd{1.0} \\
\bf{\leoauthoritative{}} & $21.3$\tablestd{0.7} & $21.2$\tablestd{0.9} & $21.1$\tablestd{0.8} & $22.3$\tablestd{0.8} & $22.9$\tablestd{0.8} & $22.1$\tablestd{0.9} & $23.2$\tablestd{0.7} & $21.9$\tablestd{0.4} & $23.9$\tablestd{1.2} & $23.0$\tablestd{1.1} & $23.1$\tablestd{0.7} \\
\bf{\leotechnical{}} & $22.5$\tablestd{0.6} & $22.4$\tablestd{0.6} & $22.5$\tablestd{0.6} & $21.2$\tablestd{0.7} & $21.8$\tablestd{0.8} & $20.5$\tablestd{0.5} & $21.1$\tablestd{0.6} & $20.5$\tablestd{0.6} & $22.1$\tablestd{0.6} & $21.2$\tablestd{0.2} & $21.4$\tablestd{0.4} \\
\bf{\leofluent{}} & $24.4$\tablestd{0.8} & $24.4$\tablestd{0.6} & $24.4$\tablestd{0.8} & $21.3$\tablestd{0.9} & $23.2$\tablestd{1.5} & $21.2$\tablestd{1.0} & $21.4$\tablestd{1.4} & $20.8$\tablestd{1.3} & $23.2$\tablestd{1.8} & $21.5$\tablestd{1.3} & $22.1$\tablestd{1.2} \\
\bf{\leociting{}} & $25.5$\tablestd{0.7} & $25.3$\tablestd{0.6} & $25.3$\tablestd{0.6} & $22.8$\tablestd{0.9} & $24.2$\tablestd{0.7} & $21.7$\tablestd{0.3} & $22.3$\tablestd{0.8} & $21.3$\tablestd{0.9} & $23.5$\tablestd{0.4} & $21.7$\tablestd{0.6} & $22.9$\tablestd{0.5} \\
\bf{\leoquotes{}} & $27.5$\tablestd{0.8} & $27.6$\tablestd{0.8} & $27.1$\tablestd{0.6} & $24.4$\tablestd{1.0} & $26.7$\tablestd{1.1} & $24.6$\tablestd{0.7} & $24.9$\tablestd{0.9} & $23.2$\tablestd{0.9} & $26.4$\tablestd{1.0} & $24.1$\tablestd{1.2} & $25.5$\tablestd{0.9} \\
\bf{\leostats{}} & $25.8$\tablestd{1.2} & $26.0$\tablestd{0.8} & $25.5$\tablestd{1.2} & $23.1$\tablestd{1.4} & $26.1$\tablestd{0.9} & $23.6$\tablestd{0.9} & $24.5$\tablestd{1.2} & $22.4$\tablestd{1.2} & $26.1$\tablestd{1.2} & $23.8$\tablestd{1.2} & $24.8$\tablestd{1.1} \\
\bottomrule
\end{tabular}
}
\caption{\label{results-app}
Absolute impression metrics of \LEO{} methods on \bench{}. Compared to baselines, simple methods like \leoseo{} traditionally used in SEO don't perform well. However, our proposed methods such as \leostats{} and \leoquotes{} show strong performance improvements across all metrics. The best methods improve upon baseline by 41\% and 28\% on \wordposmetric{} and \subjectiveimpression{} respectively.}
\label{tab:results-app}
\end{table*}

\begin{table*}[!h]
\centering
\resizebox{\linewidth}{!}{%
\begin{tabular}{lccccccccccc}
\toprule
\multirow{2}{*}[-6pt]{\large{\bf{Method}}} & \multicolumn{3}{c}{\bf{\wordposmetric}}  & \multicolumn{8}{c}{\bf{\subjectiveimpression}} \\
\cmidrule(lr){2-4} \cmidrule(lr){5-12}
& Word & Position & \bf{Overall} & Rel. & Infl. & Unique  & Div. & FollowUp & Pos. & Count & \bf{Average}
\\
\midrule
\rowcolor{gray!20} \multicolumn{12}{c}{Performance without \leo{}} \\ \midrule
\bf{\leobaseline{}} & $24.0$ & $24.4$ & $24.1$ & $24.7$ & $24.7$ & $24.7$ & $24.7$ & $24.7$ & $24.7$ & $24.7$ & $24.7$ \\
\midrule
\rowcolor{gray!20} \multicolumn{12}{c}{Non-Performing \leo{} methods} \\ \midrule
\bf{\leoseo{}} & $21.9$ & $21.4$ & $21.9$ & $26.3$ & $27.2$ & $27.2$ & $30.2$ & $27.9$ & $28.2$ & $26.9$ & $28.1$ \\
\bf{\leounique{}} & $24.0$ & $23.7$ & $23.6$ & $24.9$ & $25.1$ & $24.7$ & $24.4$ & $23.0$ & $23.6$ & $23.9$ & $24.1$ \\
\midrule
\rowcolor{gray!20} \multicolumn{12}{c}{High-Performing \leo{} methods} \\ \midrule
\bf{\leoauthoritative{}} & $25.6$ & $25.7$ & $25.9$ & $28.9$ & $30.9$ & $31.2$ & $31.7$ & $31.5$ & $26.9$ & $29.5$ & $30.6$ \\
\bf{\leofluent{}} & $25.8$ & $26.2$ & $26.0$ & $28.9$ & $29.4$ & $29.8$ & $30.6$ & $30.1$ & $29.6$ & $29.6$ & $30.0$ \\
\bf{\leociting{}} & $26.6$ & $26.9$ & $26.8$ & $19.8$ & $20.7$ & $19.5$ & $18.9$ & $20.0$ & $18.5$ & $18.9$ & $19.0$ \\
\bf{\leoquotes{}} & $28.8$ & $28.7$ & $\textbf{29.1}$ & $31.4$ & $31.9$ & $31.9$ & $32.3$ & $31.4$ & $31.7$ & $30.9$ & $32.1$ \\
\bf{\leostats{}} & $25.8$ & $26.6$ & $26.2$ & $31.6$ & $33.4$ & $34.0$ & $33.7$ & $34.0$ & $33.3$ & $33.1$ & $\textbf{33.9}$ \\
\bottomrule
\end{tabular}
}
\caption{\label{results-main-perplexity}
Performance improvement of \LEO{} methods on \bench{} with Perplexity.ai as \lee{}. 
Compared to the baselines simple methods such as \leoseo{} traditionally used in SEO often perform worse. However, our proposed methods such as \leostats{} and \leoquotes{} show strong performance improvements across the board. The best performing methods improve upon baseline by 22\% on \wordposmetric{} and 37\% on \subjectiveimpression{}. }
\label{tab:results-perplexity}
\end{table*}

\subsection{\LEO{} Methods}
\label{app:meth}

We propose 9 different \leo{} methods to optimize website content for \lee{}s. We evaluate these methods on the complete \bench{} test split. Further, to reduce variance in results, we run our experiments on five different random seeds and report the average.

\subsection{Prompts for \LEO{} methods}
\label{app:prompts}

We present all prompts in our our public repository: \url{https://github.com/GEO-optim/GEO}. GPT-3.5 turbo was used for all experiments.

\section{Results}
\label{app:results}

We perform experiments on 5 random seeds and present results with statistical deviations in Table~\ref{tab:results-app}

\subsection{\LEO{} in the Wild : Experiments with Deployed \Le{}}
\label{app:perplexity_results}

We also evaluate our proposed \leo{} methods on real-world deployed \Le{}: Perplexity.ai. Since perplexity.ai does not allow the user to specify source URLs, we instead provide source text as file uploads to perplexity.ai while ensuring all answers are generated only using the file sources provided. We evaluate all our methods on a subset of 200 samples of our test set. Results using Perplexity.ai are shown in Table~\ref{tab:results-perplexity}.

\end{document}